\documentclass[runningheads]{llncs}

\usepackage{eccv}

\usepackage{eccvabbrv}

\usepackage{graphicx}
\usepackage{booktabs}
\usepackage{multirow}
\usepackage{array,multirow,diagbox}
\usepackage{colortbl,hhline}
\newcolumntype{g}{>{\columncolor[gray]{0.85}}l}
\usepackage[accsupp]{axessibility}  
\usepackage{rotating}

\usepackage{array}
\usepackage[symbol]{footmisc}

\usepackage{hyperref}

\usepackage{orcidlink}

\begin{document}

\title{CoDA: Instructive Chain-of-Domain Adaptation with Severity-Aware Visual Prompt Tuning}

\titlerunning{CoDA}

\author{Ziyang Gong \inst{1}\thanks{These authors contributed equally.}\orcidlink{0009-0006-5191-0380}
\and Fuhao Li\inst{2}\textsuperscript{*}\orcidlink{0009-0003-4830-2171}
\and Yupeng Deng \inst{3}\textsuperscript{*}\orcidlink{0009-0008-9391-718X}
\and Deblina Bhattacharjee \inst{4}\textsuperscript{*}\orcidlink{0000-0002-0534-852X}
\and Xianzheng Ma  \inst{5}\thanks{Corresponding authors.}
\and Xiangwei Zhu \inst{1}\textsuperscript{†}\orcidlink{0000-0002-4415-0698}
\and Zhenming Ji \inst{1}\textsuperscript{†}\orcidlink{0000-0001-6786-8136}} 
\authorrunning{Gong et al.}

\institute{Sun Yat-sen University, China 
\and Wuhan University of Science and Technology, China
\and National University of Singapore, Singapore
\and EPFL, Lausanne, Switzerland
\and Independent Researcher}

\maketitle
\begin{abstract}
Unsupervised Domain Adaptation (UDA) aims to adapt models from labeled source domains to unlabeled target domains. When adapting to adverse scenes, existing UDA methods fail to perform well due to the lack of instructions, leading their models to overlook discrepancies within all adverse scenes.
To tackle this, we propose CoDA which instructs models to distinguish, focus, and learn from these discrepancies at scene and image levels.
Specifically, CoDA consists of a Chain-of-Domain (CoD) strategy and a Severity-Aware Visual Prompt Tuning (SAVPT) mechanism. CoD focuses on scene-level instructions to divide all adverse scenes into \textit{easy} and \textit{hard} scenes, guiding models to adapt from source to easy domains with easy scene images, and then to hard domains with hard scene images, thereby laying a solid foundation for whole adaptations. Building upon this foundation, we employ SAVPT to dive into more detailed image-level instructions to boost performance. SAVPT features a novel metric \textit{Severity} that divides all adverse scene images into \textit{low-severity} and \textit{high-severity} images. Then Severity directs visual prompts and adapters, instructing models to concentrate on unified severity features instead of scene-specific features, without adding complexity to the model architecture.
CoDA achieves SOTA performances on widely-used benchmarks under all adverse scenes. Notably, CoDA outperforms the existing ones by 4.6\%, and 10.3\% mIoU on the Foggy Driving, and Foggy Zurich benchmarks, respectively. Our code is available at \href{https://github.com/Cuzyoung/CoDA}{https://github.com/Cuzyoung/CoDA}. 
\end{abstract}

\section{Introduction}
\label{sec:intro}
\begin{figure}[t]
  \centering
  \includegraphics[width=\textwidth]{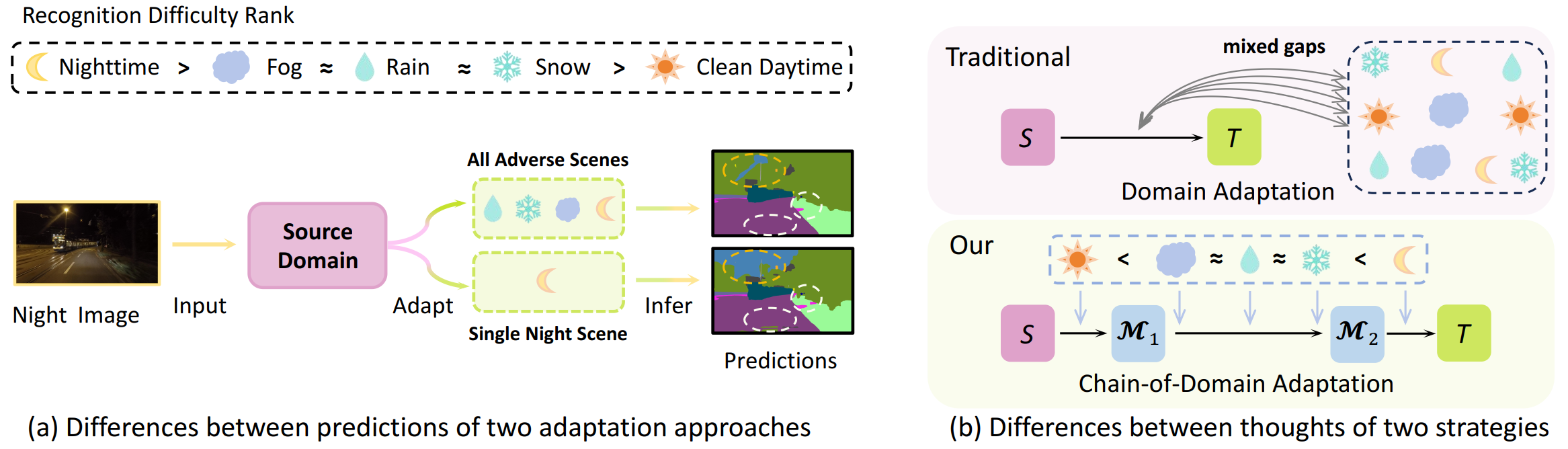}
  \caption{(a) Current SOTA models\cite{hoyer2023mic,hoyer2022hrda} trained on all adverse scenes within a target domain can achieve good performance on other details but struggle to recognize the sky under night scenes. These models, typically, 
  trained on a single night scene show the contrary results. Yellow circles in predictions denote the sky recognition and white ones indicate other classes' recognition. (b) Traditional strategy directly adapts from source to target domains with chaotic gaps. Our Chain-of-Domain (CoD) strategy \emph{instructs} models to adapt from source to target domains according to the difficulties of scenes through introducing intermediate domains.  
  }
  \label{fig: finding}
\end{figure}

Understanding all adverse scenes including fog, rain, snow, and night has become the common goal for existing UDA methods \cite{hoyer2022daformer, ma2022both, gong2023train, hoyer2023mic, li2024parsing}. Although current methods show a good understanding of fog, rain, and snow scenes, when encountering the most challenging night scene which is full of unpredictable noise due to the extremely low light conditions, their performance drops a lot. 
Our findings in Fig~\ref{fig: finding} (a) illustrate that current SOTA methods \cite{hoyer2023mic,hoyer2022hrda} train models on all adverse scenes achieve precise segmentation on details of the image shown in white circles, but they struggle to recognize the sky in night scenes shown in yellow circles. On the contrary, their models trained on a single night scene can clearly segment the sky but show worse results in other details. The results reveal that current methods are trapped in dilemmas that adapting to all adverse scenes leads models to hallucinations while adapting to a single adverse scene causes models to underfit.  

Given the essence of UDA is knowledge distillation allowing the student network to learn \textit{instructive} knowledge from the teacher network \cite{wang2021knowledge}, we argue that for existing methods, the adaptation to all adverse scenes requires scene-level instructions and the adaptation of a single adverse scene needs image-level instructions to overcome hallucinations and underfitting, respectively. Addressing these issues, we propose CoDA (\textbf{C}hain-\textbf{o}f-\textbf{D}omain \textbf{A}daptation) 
methodology focusing on these two levels to instruct and enhance models to learn domain-invariant features through the Chain-of-Domain (CoD) strategy with Severity-Aware Visual Prompt Tuning (SAVPT) mechanism. 

The design of CoD motivated by Chain-of-Thought (CoT) is focused on providing scene-level instruction. It divides all adverse scenes into \textit{easy} and \textit{hard} categories of scenes and instructs models to adapt from source to target domains, in a step-by-step fashion \cite{kojima2022large}, through extra intermediate domains as shown in Fig \ref{fig: finding} (b). As the adage says "\textit{Well begun is half done}", we maintain that acquiring high-quality prior knowledge at the start of training is critical for iterative learning. Thus, CoD starts by instructing models to train on \textit{easy} scene images to build a solid foundation, and then, CoD adapts to \textit{hard} scene images for further knowledge transfer. Additionally, as we argue that a large number of the original target images are somewhat hard for models to learn, we propose a tailored mini-dataset of adverse scenes generated by combining three Large Multimodal Models (LMMs) to serve as a part of the easy scene images for good starts to the training. More details are illustrated in the Section~\ref{generate dataset}.
In summary, CoD constructs intermediate domains according to 
scene-level difficulties of adverse scenes, instructing models to adapt to mixed gaps from easy to hard.  
 
One might ask "Why does the direct adaptation to all adverse scenes not instruct models with an easy start?" 
We denote this kind of direct adaptation as the \textit{traditional strategy} which randomly samples all adverse scene images, potentially treating disparate scenes as equivalent, overlooking their various difficulties. 
It might sample challenging night images at the onset of training causing initial errors. 
Concurrently, since pseudo-labels within UDA are inherently uncertain, the initial errors easily accumulate during iterative training. 
Thus, the traditional strategy is not suitable for models at the start of training.
Compared to CoD, however, the traditional strategy is less instructive but more diverse. Therefore, to improve scene diversity, we employ CoD to guide models during the preliminary training and activate the traditional strategy later. 

Based on the solid foundation brought by the scene-level CoD strategy, we hope to focus on a more detailed image-level perspective to further enhance the inherent abilities of models for extracting domain-invariant features. 
Since the complex architectures of existing models overfit scene-specific features, we aim to present a method that can enhance models' abilities without complicating the network architecture. The seminal work of Darcet et.al~\cite{darcet2023vision} has significantly inspired our approach, particularly their remarkable finding that \textit{visual prompts, when incorporated into the input data of Vision Transformer (ViT), can be omitted during inference, thereby enhancing the ViT's performance.} This finding validates that visual prompts can enhance models' inherent ability without being a part of network structures, which aligns well with our motivation. Thus, we present the Severity-Aware Visual Prompt Tuning (SAVPT) with an image-level metric \textit{Severity} to measure the severity differences within each image.

SAVPT contains a Severity Perception Trigger (SPT), Meta-Visual Prompts, and Meta-Adapters. SPT classifies adverse images to \textit{low-severity} and \textit{high-severity} images. The Meta-Visual Prompts and Meta-Adapters modules are learnable components that build upon the SPT mechanism and are divided into two severity branches. When encountering a low-severity image, the low-severity branch will be activated. Then the Meta-Visual Prompts and Meta-Adapters in this branch will enhance the low-severity image and optimize its features respectively. Concurrently, another branch will be frozen without training. This important mechanism ensures that the two branches will gradually demonstrate different severity emphases, leading Meta-Visual Prompts and Meta-Adapters to help models focus on severity features rather than scene-specific features. In later experiments, we will validate that SAVPT trained with models can also be discarded during inference. 

The whole CoDA method offers three key contributions: (1) CoDA is the first to propose CoT-based variants, Chain-of-Domain, in UDA for adverse scene understanding; (2) CoDA validates the findings in \cite{darcet2023vision} through experiments on the application of SAVPT during inference. This further confirms that SAVPT empowers models to learn domain-invariant features. (3) CoDA achieves state-of-the-art performance on multiple widely used adverse scene benchmarks. In particular, CoDA outperforms SOTA methods by large margins with \textbf{4.6\%} and \textbf{10.3\%} mIoU on Foggy Driving and Foggy Zurich benchmarks.

\section{Related Work}
\subsection{Semantic segmentation under adverse scenes within UDA}
Since existing works \cite{Wei_2024_CVPR, Wei_2023_ICCV, zhong2022rainy, xiao20233d, xiao2024cat} predominately focus on four adverse scenes (fog, rain, snow, and night), we divide them into three branches by the scene benchmarks that they evaluate. The first branch comprises methods focusing on parsing foggy scenes \cite{sakaridis2018model,sakaridis2018semantic, dai2020curriculum} trying to generate synthetic fog images to narrow the real-to-foggy gaps. Later works tried other paths like CuDA-Net \cite{ma2022both} that reported disentangling gaps by introducing an intermediate domain, and FIFO \cite{lee2022fifo} that presented an auxiliary network to help the segmentation model to learn fog-invariant features. Since the above works generalize well to rainy and snowy scenes, no methods are specializing in rainy and snowy scenes. Night-specialized methods \cite{wu2021dannet} are contained in the second branch and they too focus on narrowing the gaps by introducing extra domain images like twilight images or extra information\cite{dai2018dark,sakaridis2019guided,sakaridis2020map}. The final branch comprises models trying to solve multiple scenes including normal and adverse scenes. DAFormer \cite{hoyer2022daformer} was the first to introduce the SegFormer-based \cite{xie2021segformer} architecture to this field. Subsequently, some DAFormer-based methods \cite{xie2023sepico,bruggemann2023refign} emerged like STA \cite{gong2023train} which achieves domain generalization in UDA, and HRDA \cite{hoyer2022hrda} that proposes multi-scale features fusion network. Some works also take HRDA as the baseline model, such as MIC \cite{hoyer2023mic} using masks to help models learn context-level features. Different from previous methods, CoDA is the first UDA method to focus models on differences within adverse scenes to learn domain-invariant features.

\subsection{CoT and its variants}
The essential of CoT is a series of intermediate reasoning steps. CoT is proposed by \cite{wei2022chain} as a simple and effective prompting strategy to enhance the complex reasoning ability of Large Language Models (LLMs) to accomplish reasoning tasks including arithmetic, commonsense, and symbolic reasoning tasks. Recent years have witnessed an explosive growth of CoT works in LMMs \cite{ge2023chain,chen2023measuring,jacovi2024chain,himakunthala2023let,rose2023visual,mitra2023compositional}, where they arouse LLMs' and LMMs' potential and boost alignment between multimodality. Concurrently, there have been a series of works focusing on CoT variants. Specifically,
Chain-of-Thought Self-Consistency (CoT-SC) \cite{wang2022self} samples the CoT output multiple times and chooses the most consistent one, Tree-of-Thought (ToT) \cite{yao2023tree} proposes tree-structured CoT allowing LLMs to consider different reasoning paths and self-evaluate to determine the action, Graph-of-Thought (GoT) \cite{besta2023graph} creates graph-based CoT closing LLMs and human thinking, and Chain-of-Reasoning (CoR) \cite{uehara2024advancing} trains models to question themselves according to an uncertainty factor, further enhancing the answer's confidence.
However, there are no CoT variants in UDA for adverse scene understanding currently. Thus, we hope our Chain-of-Domain (CoD) method will provide an initial contribution in this field.  

\subsection{Visual prompt tuning in UDA}
Prompt-based learning \cite{liu2023pre} initially serving as Parameter-Efficient Fine-Tuning (PEFT) \cite{chen2022vision, zhang2021tip, houlsby2019parameter} is used to finetune large pre-trained models to downstream tasks in Natural Language Processing (NLP). Subsequently,
the prompt has begun to be transferred to the field of Computer Vision (CV) bringing about a significant academic trend.
Many methods \cite{radford2021learning,ju2022prompting,yao2021cpt,zhou2022learning} first tried to migrate prompt to Vision Language Models (VLM) in text form, and then VPT \cite{jia2022visual} first introduced "visual prompt" into CV as learnable vectors (soft prompt). \cite{ge2023domain} followed by other works \cite{gao2022visual,gan2023decorate,sun2023vpa} is the first work focusing on domain adaptation and employing prompt. Most relevant to our work are the methods that focus on UDA or DG for adverse scene understanding, such as \cite{fahes2023poda} employs text features to strengthen features achieving a novel style augmentation implicitly, and \cite{vidit2023clip} combines clip \cite{radford2021learning} to conduct semantic augmentation.     


\section{Approach}
\label{sec:approach}
\subsection{Preliminary}
We denote a student network as $f_\theta$ and a teacher network as $f_\phi$. Unlike traditional UDA methods, CoDA utilizes multiple domain images including source domain images $X_S \in \mathbb{R}^{H_S\times W_S\times C}$, target domain images $X_T \in \mathbb{R}^{H_T\times W_\textit{T}\times C}$, and intermediate domain images $X_{\mathcal{M}_n}  \in \mathbb{R}^{H_T\times W_T \times C}$, where $H$, $W$, and $C$ represent the height, width, and channel of images. In our experiments, we set $n$ to 2 and $X_{\mathcal{M}}$ to share the same shape with $X_T$. Thus the union of intermediate and target images is represented as $X_{T^\prime} = X_{\mathcal{M}} \cup X_T$. Notably, only source domain labels $Y_S\in\mathbb{R}^{H_S\times W_S}$ are available. All domain images are combined to calculate the Cross-Entropy (CE) loss to iteratively update $f_\theta$. Then the parameters of $f_\theta$ will be synchronized by EMA \cite{tarvainen2017mean} to $f_\phi$ which has no gradient backpropagation. 

\subsection{Training-free Pipeline to Generate Mini-Dataset}
\label{generate dataset}
Our mini-dataset contains 1200 (3$\times$400) images including fog, rain, and snow images aiming to serve as the easiest images with slightly adverse weather factors to construct a part of $M_1$. Since text prompts $\Phi$ are decisive for the quality of images generated by Stable Diffusion 2 (SD2) \cite{rombach2022high}, we propose a training-free pipeline to generate and optimize $\Phi$ step by step through collaborating with CLIP \cite{radford2021learning}, GPT-4V \cite{achiam2023gpt}, and Human Feedback. The pipeline consists of 4 stages: 

\begin{figure}[t]
    \centering
    \includegraphics[width=\textwidth]{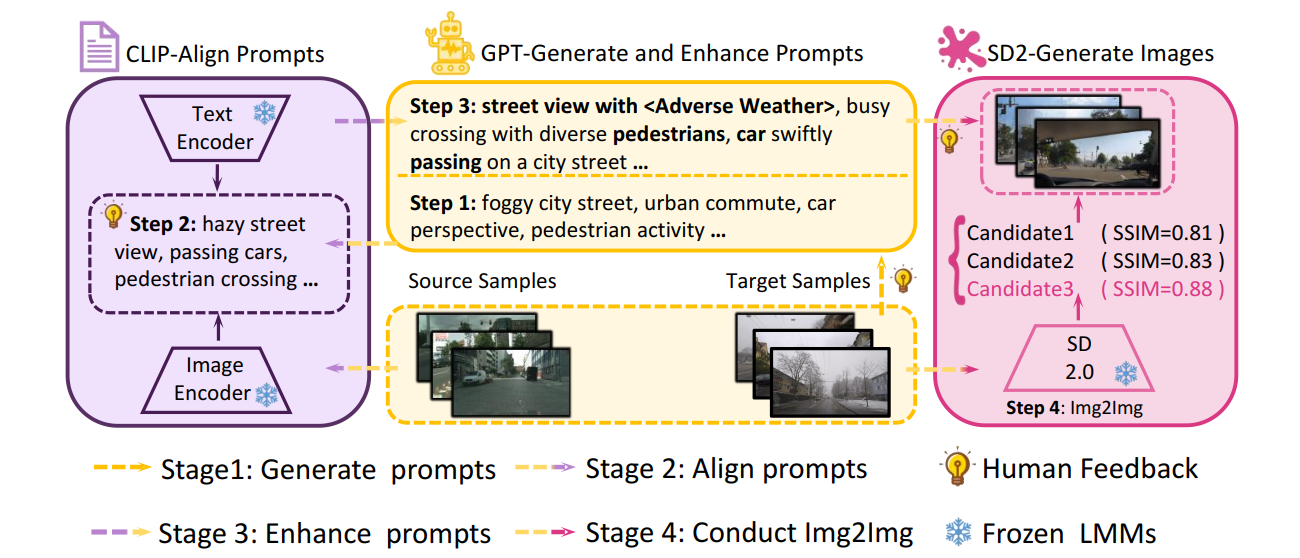}
    \caption{Four Stages Training-free Pipeline generates the mini-dataset to serve as intermediate steps providing diverse difficulty levels. All images are adverse scene images with slight weather factors generated based on ACDC adverse scene images and possess the common features of target and source images. Notably, stages 1, 2, and 4 contain manual processes based on human feedback.}
    \label{Datasets}
\end{figure}

(1) The first stage uses GPT-4V to generate the basic phrasal descriptions $\Phi_1$ of $X_T$. Notably, we aim to obtain $\Phi_1$ to represent the basic features of $X_T$. Since all $X_T$ are captured in the same city Zurich, the generated $\Phi_1$ of different images are similar to each other, except for the weather factors. Thus, we only randomly feed 6 target domain images $X_\Phi$ (fog, rain, and snow respectively contain 2 images) to GPT-4V with text instruction $\Psi_1$: "\textit{Please generate some phrases to describe the scene, style, and content of this picture, such as driving in Europe.}" Then, every image will generate dozens of output phrases like "Foggy city street", and "Urban commute" as shown in stage 1 of Fig \ref{Datasets}.
\begin{equation}
    \Phi_1 = GPT-4V(\thinspace X_\Phi, \thinspace \Psi_1 \thinspace)
\end{equation}

(2) Since the dataset is the intermediate step between source and target, the second stage aims to select the most appropriate text phrases $\Phi_2$ that represent both features of $X_S$ and $X_T$ by aligning $\Phi_1$ and $X_S$. We input $\Phi_1$ and all of $X_S$ into the CLIP text encoder $\mathcal{T}$ and image encoder $\mathcal{V}$ respectively to calculate cosine similarity. Then we select $\Phi_1$ with high similarity. 
\begin{equation}
    \Phi_2 = argmax_{C^\prime}(\frac{\mathcal{T}(\Phi_1)\cdot\mathcal{V}(X_S)}{\| \mathcal{T}(\Phi_1) \| \times \| \mathcal{V}(X_S) \|})
\end{equation}
where $C^\prime$ is the number of $\Phi_2$ for each image in $X_\Phi$ and we set it as 1 or 2. Since many phrasal prompts have similar semantic information, we manually filter the redundant based on human feedback to choose 5 texts to build $\Phi_2$: "hazy street view, passing cars, pedestrian crossing, cyclists on city lane, car Perspective." 

(3) The third stage recurs to GPT-4V to enhance $\Phi_2$. We employ GPT-4V to enrich the semantic representation of $\Phi_2$ by this instruction $\Psi_2$: "Please enrich the description of $<\Phi_2>$, requiring to maintain the simplicity and the semantic information." We denote enhanced $\Phi_2$ as $\Phi_3$: "street view with \textit{fog}, car swiftly passing on a city street, busy crossing with diverse pedestrians, cars and cyclists parked on a quiet road, inside view of a car on a busy road." 

(4) Before conducting image-to-image at stage 4, we first finetune the $\Phi_3$ based on human feedback. In detail, we adjust the <\textit{fog}> in $\Phi_3$ to "slight <fog/rain/snow>" to generate different scene images. Besides, we found that SD2 is sensitive for cyclists and it always generates a huge number of cyclists, so we changed the cyclists to motorbikes and controlled their number. Thus, the final $\Phi_4$ used to prompt SD2 consists of: "street view with \textit{slight <Adverse Weather>}, car swiftly passing on a city street, busy crossing with diverse pedestrians,  \textit{single car and motorbike} parked on a quiet road, inside view of a car on a busy road." 
We feed $\Phi_4$ and $X_T$ to SD2 to generate 3 images as candidates and choose the one with the highest SSIM \cite{wang2004image} with $X_T$ to be final images $X_G$. More details are attached to the supplementary material.

\begin{equation}
    X_G = argmax(\thinspace SSIM(\thinspace SD2(\thinspace X_T, \thinspace \Phi_4\thinspace)\thinspace)\thinspace)
\end{equation}

\subsection{Instructive Chain-of-Domain Adaptation} 
\label{CoD method}
\begin{figure}[t]
    \centering
    \includegraphics[width=\textwidth]{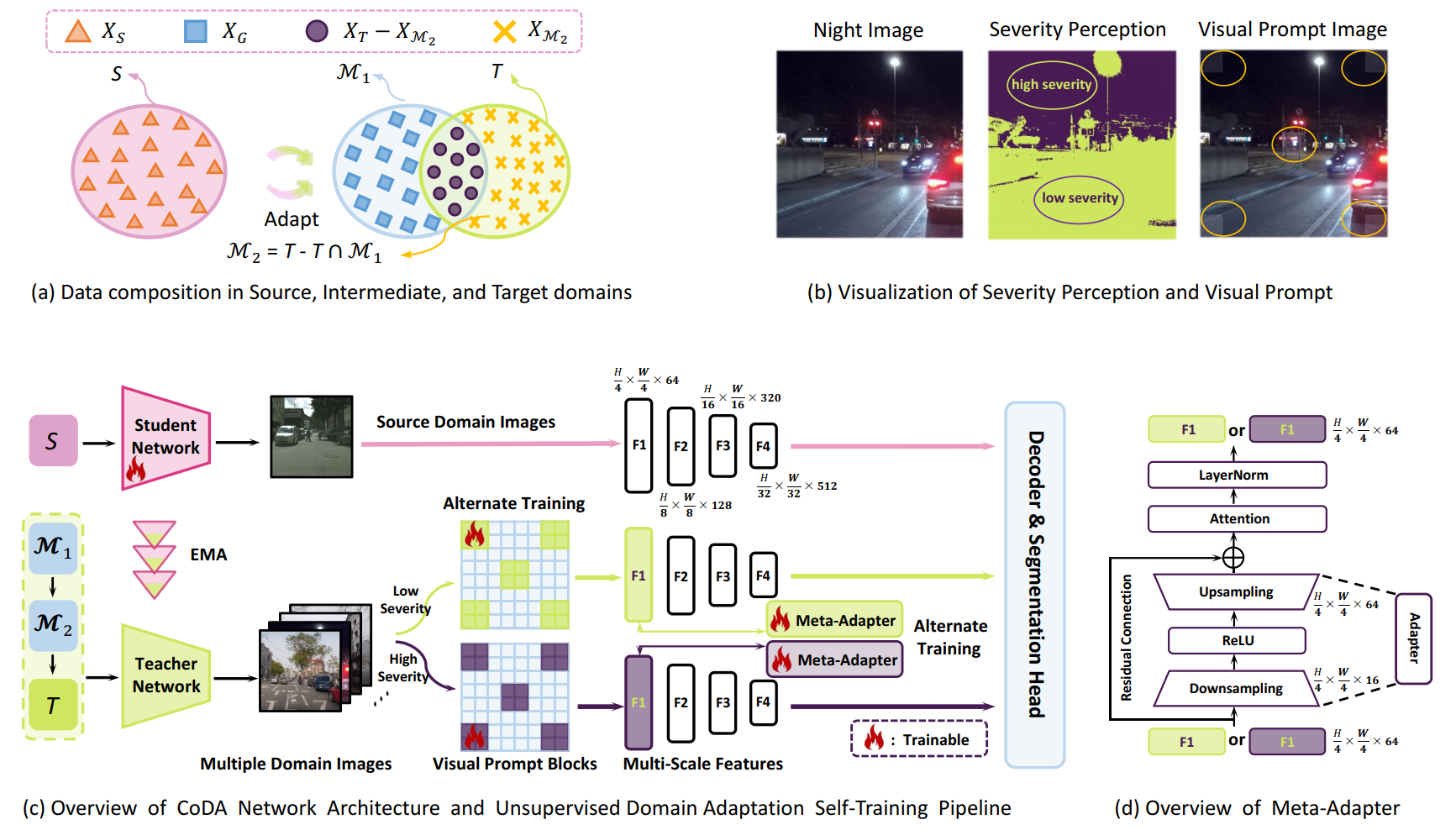}
    \caption{(a) shows the data composition in experiments of Cityscapes to ACDC. (b) demonstrates the visualization of SPT and Meta-Visual Prompts. The purple pixels are severe pixels and the green pixels are the nonsevere. (c) and (d) shows the details of CoDA's architecture and pipeline.}
    \label{Data and Arc}
\end{figure}

The first key point in scene-level CoD aims to simply divide all adverse scenes into \textbf{Easy} and \textbf{Hard} two-level scenes to construct two intermediate domains, $\mathcal{M}_1$ and $\mathcal{M}_2$. Take Cityscapes to ACDC experiment as an example, ACDC representing the target domain $T$ consists of fog, rain, snow, and night scenes, so we directly choose daytime scene images (fog, rain, snow, and $X_G$) as easy scene images to construct $X_{\mathcal{M}_1}$ and use night images to build $X_{\mathcal{M}_2}$. The data composition of Cityscapes to ACDC can be represented as Fig \ref{Data and Arc} (a). Thus, the CoD adaptation process transforms from $S \rightarrow T$ into $S \rightarrow \mathcal{M}_1 \rightarrow \mathcal{M}_2 \rightarrow T^\prime$.

The second pivot is how to train models during the CoD adaptation process. As our motivation in the Intro that "easy before difficult", we first train models on $\mathcal{M}_1$ with 1.2k iterations and then we train models on $\mathcal{M}_2$ with 4k iterations. Since $\mathcal{M}_2$ is a tougher scene than $\mathcal{M}_1$, we advocate models to spend more time on $\mathcal{M}_2$. The first 1.2k iterations can instruct good prior knowledge to models, which makes models robust enough when they first encounter tough images. Then, we repeat this process until iterations come to 12k. We also emphasize that although the traditional strategy is not suitable for the start of training, it is still advantageous as scene diversity to enhance models' performance upper bound. Thus, from 12k to the end of the training, we activate the traditional strategy to conduct random sampling from $X_{T^\prime}$ formulated as equation \ref{cod1} and \ref{cod2}.
\begin{equation}
    Iterations=\left\{\begin{array}{l}
    1.2k, \thinspace \thinspace \quad {when} \quad S \rightarrow \mathcal{M}_1 \\
    \enspace 4k, \enspace \quad {when} \quad \mathcal{M}_1 \rightarrow \mathcal{M}_2 \\
    \enspace \mathcal{1}, \enspace \quad {when} \quad \mathcal{M}_2 \rightarrow T^\prime
\end{array}\right.
    \label{cod1}
\end{equation}

\begin{equation}
    CoD=\left\{\begin{array}{l}
    S\rightarrow\mathcal{M}_1\rightarrow\mathcal{M}_2, \quad {when} \quad Iter \leqslant 12k\\
    \mathcal{M}_2 \rightarrow T^\prime, \quad until \enspace training \enspace EOF
\end{array}\right.
    \label{cod2}
\end{equation}

\subsection{Severity-Aware Visual Prompt Tuning}
SAVPT consists of a Severity Perception Trigger (SPT) mechanism, Meta-Visual Prompts, and Meta-Adapters. According to SPT, Meta-Visual Prompts, and Meta-Adapters are divided into two branches to focus on images with two severity levels. Both of these two branches share the same structure and initial parameters but train alternately. 

\textbf{Severity Perception Trigger} Different from CoD which conducts scene-level classification, the SPT mechanism focuses on image-level dividing and guides the whole SAVPT by measuring the severities of all adverse scene images. First of all, images are input to SPT will be transformed into gray-scale maps $X_g \in \mathbb{R}^{H \times W} $, and then all pixels will be divided into severe pixels and nonsevere pixels. In SPT, the pixel values lower than the gray-scale value $\sigma$ are severe pixels. If the ratio of severe pixels in an image passes the severity threshold $\tau$, this image will be classified as a high-severity image; otherwise, it will be recognized as a low-severity image. We show the visualization in Fig \ref{Data and Arc} (b) that the purple regions are severe pixels with high severity and the green pixels are nonsevere pixels with low severity.
\begin{equation}
    Severity=\left\{\begin{array}{l}
        high, \enspace if \thinspace \frac{X_g < \sigma}{H \times W} > \tau \\
        low \thinspace \thinspace, \enspace else
\end{array}\right.
    \label{e2}
\end{equation}

\textbf{Meta-Visual Prompts and Adapters} We denote Meta-Visual Prompts as $\delta_v \in \mathbb{R}^{H_v \times W_v \times C}$ which is explicitly added to the images similar to the adversarial reprogramming \cite{elsayed2018adversarial}. We set default $C=3$, $H_v=W_v=64$, and the number of $\delta_v$ for every image is 5. To avoid $\delta_v$ overly masking useful information of images, the 5 $\delta_v$ are added to the four corners and the center of images visualized in Fig \ref{Data and Arc} (b). Notably, since we aim to instruct models understanding adverse scenes, $\delta_v$ is set to be activated only on $X_{T^\prime}$ rather than $X_S$.
Thus, we denote the augmented images as ${\hat{X}_{T^\prime}}$ and augment equation as below:
\begin{equation}
    {\hat{X}_{T^\prime}} =X^H_{T^\prime}\cup X^L_{T^\prime}, \quad X^H_{T^\prime} = X^h_{T^\prime} + 5\cdot\delta^h_v, \quad X^L_{T^\prime} = X^l_{T^\prime} + 5 \cdot \delta^l_v
\end{equation}
where $X^h$, $X^l$, $\delta^h_v$, $\delta^l_v$, $X^H$, and $X^L$ respectively means that high-severity images, low-severity images, high-severity visual prompts, low-severity visual prompts, augmented $X^h$, and augmented $X^l$. The pipeline of Meta-Visual Prompts adding to images is shown in Fig \ref{Data and Arc} (c).

The design of Meta-Adapters shown in Fig \ref{Data and Arc} (d) is lightweight consisting of a vanilla adapter \cite{houlsby2019parameter} with residual connection \cite{he2016deep}, and common attention layer. The adapter layer will absorb the input feature $F_\textit{1}$ of $\hat{X}_{T^\prime}$ by downsampling and output an optimized feature by upsampling. Then, the feature will pass an attention layer $Att$ again with layer normalization $LN$ to obtain a new robust $F_\textit{1}$. During this process, Meta-Adapters aim to absorb the prompt information from Meta-Visuial Prompts. The pipeline of using Meta-Adapter can be formalized as below:
\begin{equation}
     F^{H/L}_\textit{1} = LN(\thinspace Att^{H/L}(\thinspace Adapter^{H/L}(\thinspace F_\textit{1}^{H/L} \thinspace)\thinspace)\thinspace)
\end{equation}
where $H/L$ denotes two severity branches and $F_\textit{1}$ denote one of the multi-scale features shown in Fig \ref{Data and Arc} (c) and (d).

\subsection{Loss for Training}
As we said above, only student network $f_\theta$ is trainable and the function of teacher network $f_\phi$ is generating pseudo-labels $\hat{Y}_{T^\prime}$. The whole CoDA architecture is updated by calculating CE Loss. The student loss $L_S$ as following:
\begin{equation}
     L_{S}=-\sum_{H_S,W_S}\sum_{C} Y_{S}\thinspace log f_\theta (X_{S})
\end{equation}
\begin{equation}
    \hat{Y}_{T^\prime} = [\thinspace c = argmax_{c^\prime \in C}f_\phi(X_{T^\prime})\thinspace]
\end{equation}
where $[\cdot]$ denotes the Iverson bracket. Before calculating the teacher loss $L_T$, we still need to obtain a confidence estimate $q$ \cite{hoyer2022daformer, hoyer2022hrda, hoyer2023mic} to limit the pseudo-labels which are not completely believable. Expect for $L_S$ and $L_T$, we follow our baseline setting of feature distance loss $L_{FD}$. The summary training loss function $L$ is as below:
\begin{equation}
     L_{T}=-\sum_{H_T,W_T}\sum_{C}\ q \thinspace \hat{Y}_{T^\prime} \thinspace log f_\theta(X_{T^\prime})
\end{equation}
\begin{equation}
    L = L_S + L_T + L_{FD}
\end{equation}

\section{Experiments}
\label{experiments}
\subsection{Datasets and Benchmarks}
\textbf{Cityscapes} (CS) \cite{cordts2016cityscapes} is a real-world dataset captured with normal driving scenes in 50 urban areas. \textbf{Foggy Zurich} (FZ) \cite{sakaridis2018model} and \textbf{Foggy Driving} (FD) \cite{sakaridis2018semantic} are two foggy real-world datasets. FZ contains 3808 images with light and medium fog. Notably, only 40 images are for testing in FZ. Unlike FZ, FD is a dataset purely for testing as it contains 101 labeled images.  
\textbf{Dark Zurich} (DZ) \cite{sakaridis2019guided} is captured during nighttime, twilight, and daytime including 50 validation images and 151 test images. \textbf{Nighttime Driving} (ND) \cite{dai2018dark} contains 50 nighttime images with coarsely annotated ground truth. \textbf{BDD100K-Night} (BD) \cite{yu2020bdd100k} is a subset of the BDD100K, we only use 87 test images. ND is also used for testing. \textbf{ACDC} \cite{sakaridis2021acdc} is a dataset under fog, rain, snow, and night scenes. There are 400 training images, 100 validation images (106 in night), and 500 testing images in every scene. Besides, ACDC also contains 1600 clean images called ACDC-ref.
 
\begin{table}[t]
\caption{Comparision with existing methods on all adverse scene benchmarks. Bold denotes the best mIoU, underline means the second-best mIoU, and italics denotes the performance tested by ourselves.}
\centering
\resizebox{\textwidth}{!}{
\begin{tabular}{c|c|ccc|cccc|c|c|c} 
\toprule
\multirow{3}{*}{Models}  & \multirow{3}{*}{Backbone}                              & \multicolumn{10}{c}{Adverse Scenes (mIoU)}                                                                                                                                                                       \\
                        &                               & \multicolumn{3}{c}{Foggy}                              & \multicolumn{4}{c}{Nighttime}                                              & \multicolumn{1}{c}{Rainy}     & \multicolumn{1}{c}{Snowy}     & All            \\ 
\cmidrule{3-12}
                       &                                & ACDC-Fog      & FZ            & \multicolumn{1}{c}{FD} & ACDC-Night    & DZ            & ND            & \multicolumn{1}{c}{BD} & \multicolumn{1}{c}{ACDC-Rain} & \multicolumn{1}{c}{ACDC-Snow} & ACDC-All       \\ 
\midrule
\multicolumn{12}{c}{Scene-Specialized Models}                                                                                                                                                                                                                                   \\ 
\midrule
SFSU \cite{sakaridis2018semantic}               &         RefineNet \cite{lin2017refinenet}                          & -             & 35.7          & 46.3                   & -             & -             & -             & -                      & -                             & -                             & -              \\
CMAda3 \cite{dai2020curriculum}          &          RefineNet                           & -             & 46.8          & 49.8                   & -             & -             & -             & -                      & -                             & -                             & -              \\
CuDA-Net \cite{ma2022both}          &           DeepLab-v2 \cite{chen2017deeplab}                        & 55.6          & 49.1          & 53.5          & -             & -             & -             & -                      & -                             & -                             & -              \\
FIFO \cite{lee2022fifo}       &                RefineNet                          & -             & 48.4          & 50.7                   & -             & -             & -             & -                      & -                             & -                             & -              \\
FogAdapt \cite{iqbal2022fogadapt}         &              ResNet-38                               & -             & 50.6 & 53.4                   & -             & -             & -             & -                      & -                             & -                             & -              \\
SAM-EDA \cite{wang2024exploring}          &             ViT-H \cite{dosovitskiy2020image}                              & -             & - & 56.4                   & -             & -             & -             & -                      & -                             & -                             & -              \\
GCMA \cite{sakaridis2019guided}         &       RefineNet                                  & -             & -             & -                      & -             & 42.0          & 45.6          & 33.2                   & -                             & -                             & -              \\
MCGDA  \cite{sakaridis2020map}         &        RefineNet                               & -             & -             & -                      & -             & 42.5          & 49.4          & 34.9                   & -                             & -                             & -              \\
DANNet \cite{wu2021dannet}       &          RefineNet                               & -             & -             & -                      & -             & 44.3          & 47.7          & -                      & -                             & -                             & 50.0           \\ 
\midrule
\multicolumn{12}{c}{Scene-Agnostic Models}                                                                                                                                                                                                                                  \\ 
\midrule
AdaptSeg \cite{tsai2018learning}        &       DeepLab-v2                               & -             & 26.1          & 37.6                   & -             & 30.4          & 34.5          & 22.0                   & -                             & -                             & -              \\
DAFormer \cite{hoyer2022daformer}       &        SegFormer \cite{xie2021segformer}                               & 48.9          & 40.8          & -                      & 44.7          & 48.5          & 51.8          & 33.9                   & 59.9                          & 53.7                          & 55.4           \\
SePiCo  \cite{xie2023sepico}        &              SegFormer                         & 58.5          & -             & -                      & 50.5          & 54.2          & 56.9 & 40.6         & 66.1                          & 57.9                          & 59.1           \\
STA \cite{gong2023train}    &    SegFormer    & 60.2          & 46.9             & 54.9                     & 48.4        & -          & - & -         & 61.3                         & 58.0                         & 60.9          \\
HRDA \cite{hoyer2022hrda}      &       SegFormer                                   & \underline{69.9} & 46.0          & -                      & 53.1 & 55.9          & -             & -                      & \underline{73.6}                 & \underline{69.5}                 & 68.0           \\

MIC \cite{hoyer2023mic}(baseline)          &       SegFormer                                   & \textit{67.0}             & \underline{\textit{53.3}}          & \underline{\textit{56.6}}                      & \underline{\textit{57.2}}             & \underline{60.2} & \underline{\textit{58.6}}          & \underline{\textit{41.3}}                      & \textit{72.3}                            & \textit{66.6}                             & \underline{70.4} \\ 

 \textbf{CoDA (Ours)}   &  SegFormer      & \textbf{71.8} & \textbf{60.9} & \textbf{61}          & \textbf{66.4} & \textbf{61.2} & \textbf{59.2} & \textbf{41.9}          & \textbf{75.3}                 & \textbf{70.9}                 & \textbf{72.6}  \\
\bottomrule
\end{tabular}}
\label{table1} 
\end{table}

\subsection{State-of-the-art Performance Comparison}
\textbf{Implement Details} Our performance experiments contain four aspects: foggy, nighttime, all-scenes, and other benchmarks. In experiments, we divide existing methods into two classes, the first is Scene-Specific methods and the second is Scene-Agnostic methods. For details of network settings, we follow the default settings of our baseline MIC \cite{hoyer2023mic}, including the optimizer AdamW \cite{loshchilov2017decoupled}, encoder with a learning rate of $6 \times 10^{-5}$, decoder with $6 \times 10^{-4}$ learning rate, and a linear learning rate warm-up. We mainly adjust one hyper-parameter that severity threshold $\tau$ in SPT. All experiments are conducted on a 32G Tesla V100 GPU.


\textbf{Cityscapes to foggy scenes} We conduct experiments on CS to ACDC-Fog, FZ, and FD. For the CS to ACDC-Fog, we first train CS to ACDC-All (fog, rain, snow, night) and test on ACDC-Fog. For CS to FZ and FD, we train CS to FZ and test on FZ and FD validation. The results in Table \ref{table1} show that CoDA achieves SOTA performance FogAdapt and SAM-EDA with improvements of 10.3\% and 4.6\% mIoU in FZ and FD respectively, and outperforms our baseline MIC with 4.8\% in ACDC-Fog, 7.6\%  in FZ, and 4.4\% in FD. Notably, we set $\tau$ to 0.38 in CS to ACDC and 0.05 to FZ and FD. We contend that the significant discrepancies in the values of $\tau$ can be attributed to the diverse scene composition of ACDC requires clear distinction, unlike the singular foggy scene in FZ and FD. For the data composition of CS to FZ, we distribute the light fog images in FZ as the $X_{\mathcal{M}_1}$ and medium fog images as the $X_{\mathcal{M}_2}$ without using $X_G$.

\begin{table}[t] 
  \centering
  \captionsetup{skip=1pt}
  \caption{The results of MIC and MIC tained with CoDA from CS to ACDC. w/ SAVPT indicates that SAVPT is activated during inference, while w/o SAVPT indicates the opposite. Bold means the best performance.}
  \scalebox{0.6}{
    \begin{tabular}{c|ccccccccccgcccccccc|c}
    \toprule
    Method & Road & S.walk & Build & Wall & Fence & Pole & T.Light & Sign & Veget & Terrain\newline{} & Sky & Person & Rider & Car & Truck & Bus & Train & M.bike & Bike & mIoU \\
    \midrule
    MIC & 90.8 & 67.1 & 89.2 & 54.5 & 40.5 & 57.2 & 62.0 & 68.4 & 76.3 & \textbf{61.8} & 87.0 & 71.3 & \textbf{49.4} & 89.7 & 75.7 & \textbf{86.8} & 89.1 & \textbf{56.9} & 63.0 & 70.4 \\
   w/ SAVPT & 93.1 & \textbf{72.8} & 90.7 & 57.3 & 47.4 & \textbf{59.8} & 69.8 & 69.9 & 87.2 & 59.7 & 95.4 & 71.1 & 47.3 & 90.2 & 76.7 & 82.9 & \textbf{89.8} & 55.0 & 63.7 & 72.5 \\
    w/o SAVPT & \textbf{93.1} & 72.7 & \textbf{90.7} & \textbf{57.3} & \textbf{47.4} & 56.8 & \textbf{69.9} & \textbf{70.0} & \textbf{87.3} & 59.8 & \textbf{95.4} & \textbf{71.4} & 47.6 & \textbf{90.3} & \textbf{77.1} & 83.8 & 89.1 & 54.7 & \textbf{64.1} & \textbf{72.6} \\
    \bottomrule
    \end{tabular}}
  \label{table2}%
\end{table}%
\begin{table}[t]
  \caption{Range comparison of DAFormer, HRDA. and MIC with different strategies and seeds. Bold denotes the best range and all scores are mIoU at 40k iterations. The results show that CoD-included strategies can enhance the stability of models.}
\label{cod_table}
\scriptsize
\begin{center}
\scalebox{1}{
\begin{tabular}{c|cc|ccc|c}
\toprule
Models & Strategy & Average mIoU & Seed = 1    & Seed = 2     & Seed = 38  & Range (max-min)  \\ \midrule
\multirow{3}{*}{DAFormer \cite{hoyer2022daformer}}   & Tradition        & 56.3         & 55.5 ($-0.8$) & 57.2 ($+0.9$)  & 56.3 ($+0.0$)   & 1.7      \\
                            & CoD       & 57.7          & 58.6 ($+0.9$) & 58.5 ($+0.8$)  & 56.1 ($-1.5$)  & 1.7 ($\downarrow$ 0\%) \\
                            & CoD+Tra. & 57.7          & 57.2 ($-0.5$) & 58.5 ($+0.8$)  & 57.4 ($-0.3$) & \textbf{1.3 ($\downarrow$ 24\%)} \\ \midrule
\multirow{3}{*}{HRDA \cite{hoyer2022hrda}}       & Tradition        & 64.6          & 64.8 ($+0.2$) & 62.8 ($-1.8$)  & 66.2 ($+1.6$) & 3.4  \\
                            & CoD      & 65.4          & 65.0 ($-0.4$) & 64.2 ($-1.2$) & 67.0 ($+1.6$) & 2.8 ($\downarrow $18\%) \\
                            & CoD+Tra. & 65.7          & 65.8 ($+0.1$) & 65.9 ($+0.2$)  & 65.4 ($-0.3$) & \textbf{0.5 ($\downarrow$ 85\%)} \\ \midrule
\multirow{3}{*}{MIC \cite{hoyer2023mic}} & Tradition        & 65.9          & 67.7 ($+1.8$) & 63.0 ($-2.9$)  & 67.1 ($+1.2$) & 4.7 \\
                            & CoD      & 68.3          & 67.9 ($-0.4$) & 68.4 ($+0.1$)  & 68.7 ($+0.4$) & 0.8 ($\downarrow$ 83\%)\\
                            & CoD+Tra.  & 70.0          & 70.2 ($+0.2$) & 69.6 ($-0.4$)  & 70.3 ($+0.3$) & \textbf{0.7 ($\downarrow$ 85\%)}\\ \bottomrule
\end{tabular}}
\end{center}
\end{table}   

\textbf{Cityscapes to nighttime Scenes}
The nighttime experiments contain CS to ACDC-Night, DZ, ND, and BD. Similar to the above, we also train CS to ACDC-All and test on ACDC-Night. For the rest experiments, we train CS to DZ and generalize to ND and BD. The results in Nighttime of Table \ref{table1} demonstrate that CoDA outperforms the SOTA methods with improvements of 1\% mIoU in DZ, 2.3\% in ND, and 1.3\% in BD, and outperforms MIC with 9.2\%, 0.6\%, and 0.6\% mIoU in ACDC-Night, ND, and BD respectively. From CS to DZ, ND, and BD, we designate the threshold $\tau$ of 0.05. Additionally, we allocated 400 night images in the ACDC-ref to establish $\mathcal{M}_1$, while utilizing the original night images from DZ to form $\mathcal{M}_2$.

\textbf{Cityscapes to rainy, snowy, and all scenes} We conduct CS to ACDC-All and test on ACDC-Rain, Snow, and All for these experiments. As shown in Table \ref{table1}, CoDA achieves 72.6\% mIoU on the ACDC-All benchmark outperforming the MIC 2.2\%, and demonstrates strong generalizability on ACDC-Rain and Snow with improvements of 3.0\% and 4.3\% respectively compared to MIC. For more details, we show the results of every class of CS to ACDC-All in Table \ref{table2}. As illustrated in the Intro section, the previous methods show weakness in recognizing the sky under night scenes, which our CoDA alleviates with 8.4\% improvements in the sky class compared to MIC. 

\begin{figure}[t]
    \centering
    \includegraphics[width=\textwidth]{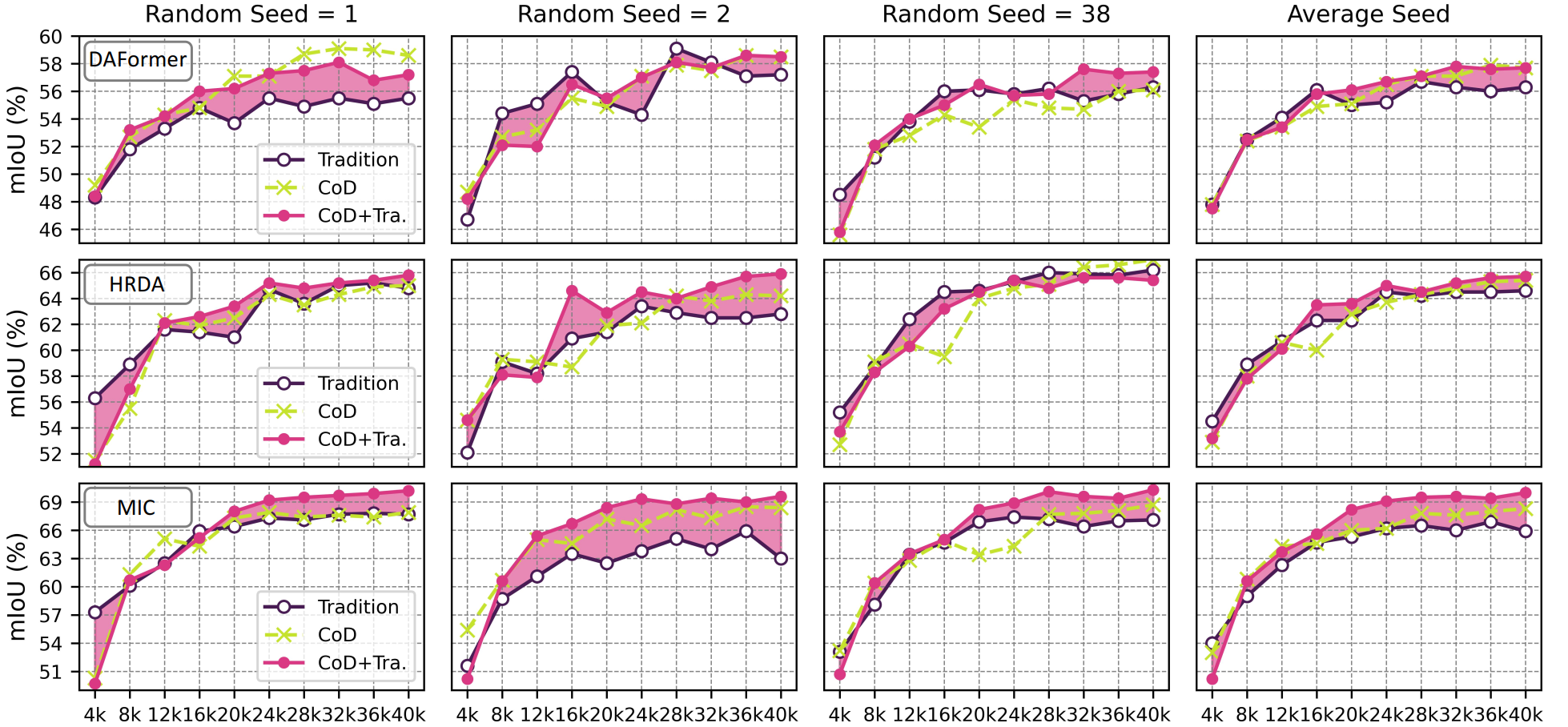}
    \caption{Ablation studies on CS to ACDC val. The x and y axes respectively mean the mIoU and Iteration. The purple, green, and red lines respectively mean original models with traditional strategy, CoD strategy, and CoD$+$traditional strategy that we implement in CoDA.}
    \label{cod ablation}
\end{figure}

\subsection{Ablation Study Analysis}
\textbf{Well begun enhances models stability} We defend that CoD is instructive for models to acquire high-quality prior knowledge which is critical for models' robustness. To validate it, we conduct CS to ACDC (val) experiments on 3 SegFormer \cite{xie2021segformer}-based UDA models, DAFormer \cite{hoyer2022daformer}, HRDA \cite{hoyer2022hrda}, and MIC \cite{hoyer2023mic} with 3 random seeds shown in Table \ref{cod_table}. Every model is trained with three different strategies, traditional strategy, Chain-of-Domain (CoD), and Chain-of-Domain+traditional strategy (CoD+Tra.).

Notably, we mainly compare the \textit{Range} in Table \ref{cod_table} to show the stability of models. We first focus on the range of models with traditional strategy. The ranges escalate from 1.7 of DAFormer to 4.1 of MIC which means that with the performance and the complexity of models improving, the instability is worse. CoD can alleviate it by stabilizing the ranges with improvements of 18\% range in HRDA and 83\% range in MIC. The results also show that the range of DAFormer with CoD is not optimized. We argue the reason is that DAFormer with low complexity is relatively stable. But CoD optimizes DAFormer performances with 1.4\% improvements in the average mIoU. Notably, CoD with the traditional strategy brings huge improvement to every model with 24\%, 85\%, and 85\% ranges in DAFormer, HRDA, and MIC respectively, which means that the traditional strategy possessing diversity can also stabilize the models when it is employed with CoD. To observe the tendency of Table \ref{cod_table}, we also show the line charts of performance from 0 to 40k iterations in Fig \ref{cod ablation}. The red shadows demonstrate models trained CoD with the traditional strategy outperform the single traditional strategy significantly.

\renewcommand{\arraystretch}{1.2} 
 \begin{table}[t]
\centering
\scriptsize
\caption{Ablation studies of positions and initialization of Meta-Visual Prompts on CS to ACDC val. All the mIoU score is the highest score during the training. $\delta_v \sim \mathcal{U}(0,1)$ and $\delta_v \sim \mathcal{N}(0,1)$ mean $\delta_v$ samples from uniform and normal distributions respectively. }
\resizebox{0.8\linewidth}{!}{
\begin{tabular}{c|c|c|c|c}
\toprule
    \multirow{2}{*}{Position of Prompts} & \multicolumn{4}{c}{Initial Parameter} \\
    \cmidrule{2-5} & 
    $\quad$ $\delta_v=1$ $\quad$ & $\quad$ $\delta_v \sim \mathcal{U}(0,1)$ $\quad$ & $\quad$ $\delta_v \sim \mathcal{N}(0,1)$ $\quad$ & $\quad$ $\delta_v =0$ $\quad$  \\ 
     \midrule
     \multicolumn{5}{c}{Best mIoU results during 0$\sim$40k iteration} \\ 
  \midrule
   Random Patch& \textbf{70.8} & 68.6  & 69.0 & 71.1 \\ 
  \midrule
   Padding Patch & 67.2  & \textbf{69.4}& \textbf{69.9} & \textbf{71.4} \\ 
  \midrule
   Corner$+$Center & 67.6 & \textbf{69.4} & 69.2 &  71.3   \\ 
   \midrule
   \multicolumn{5}{c}{Best mIoU results during 40$\sim$60k iteration}\\
   \midrule
   Random Patch& \textbf{70.8} & 69.6  & \textbf{70.2} & 71.8 \\ 
  \midrule
   Padding Patch & 67.4  & 69.5 & 69.2 $\downarrow$& 71.6 \\ 
  \midrule
   Corner$+$Center & 67.7 & \textbf{69.7} & 69.6 &  \textbf{72.1}   \\ 
\bottomrule
\end{tabular}}
\label{table5}
\end{table}

\begin{figure}[t]
    \centering
    \includegraphics[width=\textwidth]{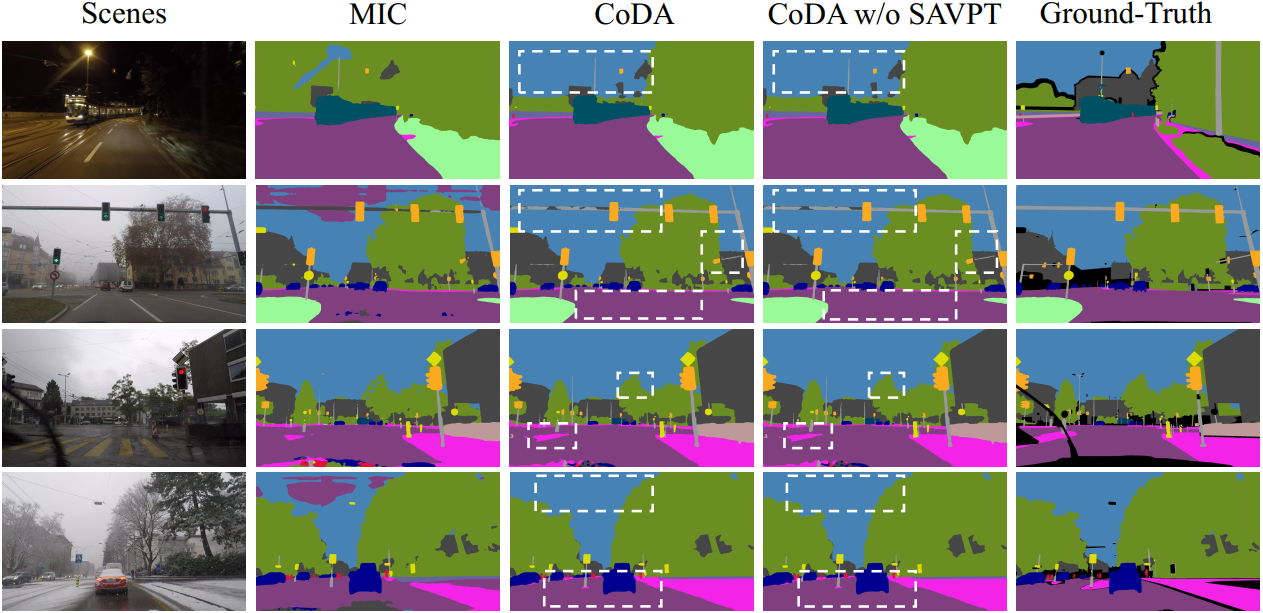}
    \caption{Quantitative experiments between MIC, MIC trained with CoDA, and Mic trained with CoDA but without SAVPT during inference time. The results reveal that CoDA understands all scenes better and SAVPT enhances models' abilities.}
    \label{SAVPT}
\end{figure}

\textbf{SAVPT enhances models' inherent abilities} 
We create SAVPT to enhance models' inherent abilities to learn domain-invariant features without complicating original models' architectures. To validate it, we conduct ablation studies of the activation of SAVPT during inference time. As illustrated in Table \ref{table2}, inference w/ SAVPT and w/o SAVPT achieve similar performance, and w/o activating SAVPT even shows superiority over w/ SAVPT by 0.1\% mIoU, which shows SAVPT indeed strengthens the models themselves rather than serve as a part of models. Concurrently, compared to the original MIC model, MIC trained with CoDA achieves obvious improvement in road, sidewalk, wall, fence, vegetation, and sky classes which empirically represent domain-invariant classes shown in Fig \ref{SAVPT}. Thus, these phenomena effectively validate that SAVPT is a knock-down composition that can lead models to learn domain-invariant features, which aligns well with our motivation. Remarkably, our findings based on \cite{darcet2023vision} unveil another interesting phenomenon: not only can visual prompts be discarded at the inference time without negatively impacting models' performance, but adapters also exhibit attributes akin to that of visual prompts. This observation can probably be ascribed to the efficient yet potent design of both the original adapters and our Meta-Adapters, allowing the implementation of SAVPT without requiring additional computational costs during inference.

Although the adding position of visual prompts is discussed in \cite{bahng2022exploring, gan2023decorate, zhang2024instruct}, the discussion of the initial parameters of visual prompts is relatively blank. Thus, we conduct joint experiments of positions with different initializations shown in Table \ref{table5}. During 40$\sim$60k, almost each experiment improves compared to their 0$\sim$40k scores and the Corner+Center shows the best performance with 72.1\% mIoU. When initial $\delta_v$=1, Padding Patch and Corner+Center performances reduce a lot contrary to Random Patch with 70.8\% mIoU. We argue that $\delta_v$=1 gives models a noisy begin and enough randomness can alleviate the noise influences. Also, the initialization with uniform and normal distribution achieves similar mIoU, which means MIC is not sensitive to initial parameter distributions and results of $\delta_v$=0 demonstrate that starting from scratch is the best choice for Meta-Visual Prompts.

\section{Conclusion}

In this paper, we propose a method named CoDA, which consists of a CoD strategy with a tailored dataset and SAVPT mechanism. Extensive experiments validate that CoD provides scene-level instructions to models for eliminating hallucinations on tough scenes, while SAVPT serves as a plug-in mechanism enhancing models' inherent abilities without complicating networks through image-level instructions. In this paper, however, we do not detailedly discuss why the Meta-adapter emerges attributes akin to visual prompts that can be discarded during training. This important phenomenon will be one of our future investigations. 

\section*{Acknowledgement}
This study was funded by the National Natural Science Foundation of China (Grants No. U21A6001) and the Innovation Group Project of Southern Marine Science and Engineering Guangdong Laboratory (Zhuhai) (Grants No. $\quad$ SML2023SP208). We also acknowledge the high-performance computing support from School of Atmospheric Science at Sun Yat-sen University.

%
%
\bibliographystyle{splncs04}
\bibliography{main}

\begin{thebibliography}{10}
\providecommand{\url}[1]{\texttt{#1}}
\providecommand{\urlprefix}{URL }
\providecommand{\doi}[1]{https://doi.org/#1}

\bibitem{achiam2023gpt}
Achiam, J., Adler, S., Agarwal, S., Ahmad, L., Akkaya, I., Aleman, F.L., Almeida, D., Altenschmidt, J., Altman, S., Anadkat, S., et~al.: Gpt-4 technical report. arXiv preprint arXiv:2303.08774  (2023)

\bibitem{bahng2022exploring}
Bahng, H., Jahanian, A., Sankaranarayanan, S., Isola, P.: Exploring visual prompts for adapting large-scale models. arXiv preprint arXiv:2203.17274  (2022)

\bibitem{besta2023graph}
Besta, M., Blach, N., Kubicek, A., Gerstenberger, R., Gianinazzi, L., Gajda, J., Lehmann, T., Podstawski, M., Niewiadomski, H., Nyczyk, P., et~al.: Graph of thoughts: Solving elaborate problems with large language models. arXiv preprint arXiv:2308.09687  (2023)

\bibitem{bruggemann2023refign}
Br{\"u}ggemann, D., Sakaridis, C., Truong, P., Van~Gool, L.: Refign: Align and refine for adaptation of semantic segmentation to adverse conditions. In: Proceedings of the IEEE/CVF Winter Conference on Applications of Computer Vision. pp. 3174--3184 (2023)

\bibitem{chen2017deeplab}
Chen, L.C., Papandreou, G., Kokkinos, I., Murphy, K., Yuille, A.L.: Deeplab: Semantic image segmentation with deep convolutional nets, atrous convolution, and fully connected crfs. IEEE transactions on pattern analysis and machine intelligence  \textbf{40}(4),  834--848 (2017)

\bibitem{chen2023measuring}
Chen, Y., Sikka, K., Cogswell, M., Ji, H., Divakaran, A.: Measuring and improving chain-of-thought reasoning in vision-language models. arXiv preprint arXiv:2309.04461  (2023)

\bibitem{chen2022vision}
Chen, Z., Duan, Y., Wang, W., He, J., Lu, T., Dai, J., Qiao, Y.: Vision transformer adapter for dense predictions. arXiv preprint arXiv:2205.08534  (2022)

\bibitem{cordts2016cityscapes}
Cordts, M., Omran, M., Ramos, S., Rehfeld, T., Enzweiler, M., Benenson, R., Franke, U., Roth, S., Schiele, B.: The cityscapes dataset for semantic urban scene understanding. In: Proceedings of the IEEE conference on computer vision and pattern recognition. pp. 3213--3223 (2016)

\bibitem{dai2020curriculum}
Dai, D., Sakaridis, C., Hecker, S., Van~Gool, L.: Curriculum model adaptation with synthetic and real data for semantic foggy scene understanding. International Journal of Computer Vision  \textbf{128},  1182--1204 (2020)

\bibitem{dai2018dark}
Dai, D., Van~Gool, L.: Dark model adaptation: Semantic image segmentation from daytime to nighttime. In: 2018 21st International Conference on Intelligent Transportation Systems (ITSC). pp. 3819--3824. IEEE (2018)

\bibitem{darcet2023vision}
Darcet, T., Oquab, M., Mairal, J., Bojanowski, P.: Vision transformers need registers. arXiv preprint arXiv:2309.16588  (2023)

\bibitem{dosovitskiy2020image}
Dosovitskiy, A., Beyer, L., Kolesnikov, A., Weissenborn, D., Zhai, X., Unterthiner, T., Dehghani, M., Minderer, M., Heigold, G., Gelly, S., et~al.: An image is worth 16x16 words: Transformers for image recognition at scale. arXiv preprint arXiv:2010.11929  (2020)

\bibitem{elsayed2018adversarial}
Elsayed, G.F., Goodfellow, I., Sohl-Dickstein, J.: Adversarial reprogramming of neural networks. arXiv preprint arXiv:1806.11146  (2018)

\bibitem{fahes2023poda}
Fahes, M., Vu, T.H., Bursuc, A., P{\'e}rez, P., De~Charette, R.: Poda: Prompt-driven zero-shot domain adaptation. In: Proceedings of the IEEE/CVF International Conference on Computer Vision. pp. 18623--18633 (2023)

\bibitem{gan2023decorate}
Gan, Y., Bai, Y., Lou, Y., Ma, X., Zhang, R., Shi, N., Luo, L.: Decorate the newcomers: Visual domain prompt for continual test time adaptation. In: Proceedings of the AAAI Conference on Artificial Intelligence. vol.~37, pp. 7595--7603 (2023)

\bibitem{gao2022visual}
Gao, Y., Shi, X., Zhu, Y., Wang, H., Tang, Z., Zhou, X., Li, M., Metaxas, D.N.: Visual prompt tuning for test-time domain adaptation. arXiv preprint arXiv:2210.04831  (2022)

\bibitem{ge2023domain}
Ge, C., Huang, R., Xie, M., Lai, Z., Song, S., Li, S., Huang, G.: Domain adaptation via prompt learning. IEEE Transactions on Neural Networks and Learning Systems  (2023)

\bibitem{ge2023chain}
Ge, J., Luo, H., Qian, S., Gan, Y., Fu, J., Zhan, S.: Chain of thought prompt tuning in vision language models. arXiv preprint arXiv:2304.07919  (2023)

\bibitem{gong2023train}
Gong, Z., Li, F., Deng, Y., Shen, W., Ma, X., Ji, Z., Xia, N.: Train one, generalize to all: Generalizable semantic segmentation from single-scene to all adverse scenes. In: Proceedings of the 31st ACM International Conference on Multimedia. pp. 2275--2284 (2023)

\bibitem{he2016deep}
He, K., Zhang, X., Ren, S., Sun, J.: Deep residual learning for image recognition. In: Proceedings of the IEEE conference on computer vision and pattern recognition. pp. 770--778 (2016)

\bibitem{himakunthala2023let}
Himakunthala, V., Ouyang, A., Rose, D., He, R., Mei, A., Lu, Y., Sonar, C., Saxon, M., Wang, W.: Let’s think frame by frame with vip: A video infilling and prediction dataset for evaluating video chain-of-thought. In: Proceedings of the 2023 Conference on Empirical Methods in Natural Language Processing. pp. 204--219 (2023)

\bibitem{houlsby2019parameter}
Houlsby, N., Giurgiu, A., Jastrzebski, S., Morrone, B., De~Laroussilhe, Q., Gesmundo, A., Attariyan, M., Gelly, S.: Parameter-efficient transfer learning for nlp. In: International Conference on Machine Learning. pp. 2790--2799. PMLR (2019)

\bibitem{hoyer2022daformer}
Hoyer, L., Dai, D., Van~Gool, L.: Daformer: Improving network architectures and training strategies for domain-adaptive semantic segmentation. In: Proceedings of the IEEE/CVF Conference on Computer Vision and Pattern Recognition. pp. 9924--9935 (2022)

\bibitem{hoyer2022hrda}
Hoyer, L., Dai, D., Van~Gool, L.: Hrda: Context-aware high-resolution domain-adaptive semantic segmentation. In: European Conference on Computer Vision. pp. 372--391. Springer (2022)

\bibitem{hoyer2023mic}
Hoyer, L., Dai, D., Wang, H., Van~Gool, L.: Mic: Masked image consistency for context-enhanced domain adaptation. In: Proceedings of the IEEE/CVF Conference on Computer Vision and Pattern Recognition. pp. 11721--11732 (2023)

\bibitem{iqbal2022fogadapt}
Iqbal, J., Hafiz, R., Ali, M.: Fogadapt: Self-supervised domain adaptation for semantic segmentation of foggy images. Neurocomputing  \textbf{501},  844--856 (2022)

\bibitem{jacovi2024chain}
Jacovi, A., Bitton, Y., Bohnet, B., Herzig, J., Honovich, O., Tseng, M., Collins, M., Aharoni, R., Geva, M.: A chain-of-thought is as strong as its weakest link: A benchmark for verifiers of reasoning chains. arXiv preprint arXiv:2402.00559  (2024)

\bibitem{jia2022visual}
Jia, M., Tang, L., Chen, B.C., Cardie, C., Belongie, S., Hariharan, B., Lim, S.N.: Visual prompt tuning. In: European Conference on Computer Vision. pp. 709--727. Springer (2022)

\bibitem{ju2022prompting}
Ju, C., Han, T., Zheng, K., Zhang, Y., Xie, W.: Prompting visual-language models for efficient video understanding. In: European Conference on Computer Vision. pp. 105--124. Springer (2022)

\bibitem{kojima2022large}
Kojima, T., Gu, S.S., Reid, M., Matsuo, Y., Iwasawa, Y.: Large language models are zero-shot reasoners. Advances in neural information processing systems  \textbf{35},  22199--22213 (2022)

\bibitem{lee2022fifo}
Lee, S., Son, T., Kwak, S.: Fifo: Learning fog-invariant features for foggy scene segmentation. In: Proceedings of the IEEE/CVF Conference on Computer Vision and Pattern Recognition. pp. 18911--18921 (2022)

\bibitem{li2024parsing}
Li, F., Gong, Z., Deng, Y., Ma, X., Zhang, R., Ji, Z., Zhu, X., Zhang, H.: Parsing all adverse scenes: Severity-aware semantic segmentation with mask-enhanced cross-domain consistency. In: Proceedings of the AAAI Conference on Artificial Intelligence. vol.~38, pp. 13483--13491 (2024)

\bibitem{lin2017refinenet}
Lin, G., Milan, A., Shen, C., Reid, I.: Refinenet: Multi-path refinement networks for high-resolution semantic segmentation. In: Proceedings of the IEEE conference on computer vision and pattern recognition. pp. 1925--1934 (2017)

\bibitem{liu2023pre}
Liu, P., Yuan, W., Fu, J., Jiang, Z., Hayashi, H., Neubig, G.: Pre-train, prompt, and predict: A systematic survey of prompting methods in natural language processing. ACM Computing Surveys  \textbf{55}(9),  1--35 (2023)

\bibitem{loshchilov2017decoupled}
Loshchilov, I., Hutter, F.: Decoupled weight decay regularization. arXiv preprint arXiv:1711.05101  (2017)

\bibitem{ma2022both}
Ma, X., Wang, Z., Zhan, Y., Zheng, Y., Wang, Z., Dai, D., Lin, C.W.: Both style and fog matter: Cumulative domain adaptation for semantic foggy scene understanding. In: Proceedings of the IEEE/CVF Conference on Computer Vision and Pattern Recognition. pp. 18922--18931 (2022)

\bibitem{mitra2023compositional}
Mitra, C., Huang, B., Darrell, T., Herzig, R.: Compositional chain-of-thought prompting for large multimodal models. arXiv preprint arXiv:2311.17076  (2023)

\bibitem{radford2021learning}
Radford, A., Kim, J.W., Hallacy, C., Ramesh, A., Goh, G., Agarwal, S., Sastry, G., Askell, A., Mishkin, P., Clark, J., et~al.: Learning transferable visual models from natural language supervision. In: International conference on machine learning. pp. 8748--8763. PMLR (2021)

\bibitem{richter2016playing}
Richter, S.R., Vineet, V., Roth, S., Koltun, V.: Playing for data: Ground truth from computer games. In: Computer Vision--ECCV 2016: 14th European Conference, Amsterdam, The Netherlands, October 11-14, 2016, Proceedings, Part II 14. pp. 102--118. Springer (2016)

\bibitem{rombach2022high}
Rombach, R., Blattmann, A., Lorenz, D., Esser, P., Ommer, B.: High-resolution image synthesis with latent diffusion models. In: Proceedings of the IEEE/CVF conference on computer vision and pattern recognition. pp. 10684--10695 (2022)

\bibitem{rose2023visual}
Rose, D., Himakunthala, V., Ouyang, A., He, R., Mei, A., Lu, Y., Saxon, M., Sonar, C., Mirza, D., Wang, W.Y.: Visual chain of thought: Bridging logical gaps with multimodal infillings. arXiv preprint arXiv:2305.02317  (2023)

\bibitem{sakaridis2019guided}
Sakaridis, C., Dai, D., Gool, L.V.: Guided curriculum model adaptation and uncertainty-aware evaluation for semantic nighttime image segmentation. In: Proceedings of the IEEE/CVF International Conference on Computer Vision. pp. 7374--7383 (2019)

\bibitem{sakaridis2018model}
Sakaridis, C., Dai, D., Hecker, S., Van~Gool, L.: Model adaptation with synthetic and real data for semantic dense foggy scene understanding. In: Proceedings of the european conference on computer vision (ECCV). pp. 687--704 (2018)

\bibitem{sakaridis2018semantic}
Sakaridis, C., Dai, D., Van~Gool, L.: Semantic foggy scene understanding with synthetic data. International Journal of Computer Vision  \textbf{126},  973--992 (2018)

\bibitem{sakaridis2020map}
Sakaridis, C., Dai, D., Van~Gool, L.: Map-guided curriculum domain adaptation and uncertainty-aware evaluation for semantic nighttime image segmentation. IEEE Transactions on Pattern Analysis and Machine Intelligence  \textbf{44}(6),  3139--3153 (2020)

\bibitem{sakaridis2021acdc}
Sakaridis, C., Dai, D., Van~Gool, L.: Acdc: The adverse conditions dataset with correspondences for semantic driving scene understanding. In: Proceedings of the IEEE/CVF International Conference on Computer Vision. pp. 10765--10775 (2021)

\bibitem{sun2023vpa}
Sun, J., Ibrahim, M., Hall, M., Evtimov, I., Mao, Z.M., Ferrer, C.C., Hazirbas, C.: Vpa: Fully test-time visual prompt adaptation. In: Proceedings of the 31st ACM International Conference on Multimedia. pp. 5796--5806 (2023)

\bibitem{tarvainen2017mean}
Tarvainen, A., Valpola, H.: Mean teachers are better role models: Weight-averaged consistency targets improve semi-supervised deep learning results. Advances in neural information processing systems  \textbf{30} (2017)

\bibitem{tsai2018learning}
Tsai, Y.H., Hung, W.C., Schulter, S., Sohn, K., Yang, M.H., Chandraker, M.: Learning to adapt structured output space for semantic segmentation. In: Proceedings of the IEEE conference on computer vision and pattern recognition. pp. 7472--7481 (2018)

\bibitem{uehara2024advancing}
Uehara, K., Goswami, N., Wang, H., Baba, T., Tanaka, K., Hashimoto, T., Wang, K., Ito, R., Naoya, T., Umagami, R., et~al.: Advancing large multi-modal models with explicit chain-of-reasoning and visual question generation. arXiv preprint arXiv:2401.10005  (2024)

\bibitem{vidit2023clip}
Vidit, V., Engilberge, M., Salzmann, M.: Clip the gap: A single domain generalization approach for object detection. In: Proceedings of the IEEE/CVF Conference on Computer Vision and Pattern Recognition. pp. 3219--3229 (2023)

\bibitem{wang2021knowledge}
Wang, L., Yoon, K.J.: Knowledge distillation and student-teacher learning for visual intelligence: A review and new outlooks. IEEE transactions on pattern analysis and machine intelligence  \textbf{44}(6),  3048--3068 (2021)

\bibitem{wang2022self}
Wang, X., Wei, J., Schuurmans, D., Le, Q., Chi, E., Narang, S., Chowdhery, A., Zhou, D.: Self-consistency improves chain of thought reasoning in language models. arXiv preprint arXiv:2203.11171  (2022)

\bibitem{wang2004image}
Wang, Z., Bovik, A.C., Sheikh, H.R., Simoncelli, E.P.: Image quality assessment: from error visibility to structural similarity. IEEE transactions on image processing  \textbf{13}(4),  600--612 (2004)

\bibitem{wang2024exploring}
Wang, Z., Zhang, Y., Zhang, Z., Jiang, Z., Yu, Y., Li, L., Li, L.: Exploring semantic prompts in the segment anything model for domain adaptation. Remote Sensing  \textbf{16}(5), ~758 (2024)

\bibitem{wei2022chain}
Wei, J., Wang, X., Schuurmans, D., Bosma, M., Xia, F., Chi, E., Le, Q.V., Zhou, D., et~al.: Chain-of-thought prompting elicits reasoning in large language models. Advances in Neural Information Processing Systems  \textbf{35},  24824--24837 (2022)

\bibitem{Wei_2024_CVPR}
Wei, Z., Chen, L., Jin, Y., Ma, X., Liu, T., Ling, P., Wang, B., Chen, H., Zheng, J.: Stronger fewer \& superior: Harnessing vision foundation models for domain generalized semantic segmentation. In: Proceedings of the IEEE/CVF Conference on Computer Vision and Pattern Recognition (CVPR). pp. 28619--28630 (June 2024)

\bibitem{Wei_2023_ICCV}
Wei, Z., Chen, L., Tu, T., Ling, P., Chen, H., Jin, Y.: Disentangle then parse: Night-time semantic segmentation with illumination disentanglement. In: Proceedings of the IEEE/CVF International Conference on Computer Vision (ICCV). pp. 21593--21603 (October 2023)

\bibitem{wu2021dannet}
Wu, X., Wu, Z., Guo, H., Ju, L., Wang, S.: Dannet: A one-stage domain adaptation network for unsupervised nighttime semantic segmentation. In: Proceedings of the IEEE/CVF Conference on Computer Vision and Pattern Recognition. pp. 15769--15778 (2021)

\bibitem{xiao20233d}
Xiao, A., Huang, J., Xuan, W., Ren, R., Liu, K., Guan, D., El~Saddik, A., Lu, S., Xing, E.P.: 3d semantic segmentation in the wild: Learning generalized models for adverse-condition point clouds. In: Proceedings of the IEEE/CVF Conference on Computer Vision and Pattern Recognition. pp. 9382--9392 (2023)

\bibitem{xiao2024cat}
Xiao, A., Xuan, W., Qi, H., Xing, Y., Ren, R., Zhang, X., Lu, S.: Cat-sam: Conditional tuning network for few-shot adaptation of segmentation anything model. arXiv preprint arXiv:2402.03631  (2024)

\bibitem{xie2023sepico}
Xie, B., Li, S., Li, M., Liu, C.H., Huang, G., Wang, G.: Sepico: Semantic-guided pixel contrast for domain adaptive semantic segmentation. IEEE Transactions on Pattern Analysis and Machine Intelligence  (2023)

\bibitem{xie2021segformer}
Xie, E., Wang, W., Yu, Z., Anandkumar, A., Alvarez, J.M., Luo, P.: Segformer: Simple and efficient design for semantic segmentation with transformers. Advances in Neural Information Processing Systems  \textbf{34},  12077--12090 (2021)

\bibitem{yao2023tree}
Yao, S., Yu, D., Zhao, J., Shafran, I., Griffiths, T.L., Cao, Y., Narasimhan, K.: Tree of thoughts: Deliberate problem solving with large language models. arXiv preprint arXiv:2305.10601  (2023)

\bibitem{yao2021cpt}
Yao, Y., Zhang, A., Zhang, Z., Liu, Z., Chua, T.S., Sun, M.: Cpt: Colorful prompt tuning for pre-trained vision-language models. arXiv preprint arXiv:2109.11797  (2021)

\bibitem{yu2020bdd100k}
Yu, F., Chen, H., Wang, X., Xian, W., Chen, Y., Liu, F., Madhavan, V., Darrell, T.: Bdd100k: A diverse driving dataset for heterogeneous multitask learning. In: Proceedings of the IEEE/CVF conference on computer vision and pattern recognition. pp. 2636--2645 (2020)

\bibitem{zhang2024instruct}
Zhang, J., Wang, B., Li, L., Nakashima, Y., Nagahara, H.: Instruct me more! random prompting for visual in-context learning. In: Proceedings of the IEEE/CVF Winter Conference on Applications of Computer Vision. pp. 2597--2606 (2024)

\bibitem{zhang2021tip}
Zhang, R., Fang, R., Zhang, W., Gao, P., Li, K., Dai, J., Qiao, Y., Li, H.: Tip-adapter: Training-free clip-adapter for better vision-language modeling. arXiv preprint arXiv:2111.03930  (2021)

\bibitem{zhong2022rainy}
Zhong, X., Tu, S., Ma, X., Jiang, K., Huang, W., Wang, Z.: Rainy wcity: A real rainfall dataset with diverse conditions for semantic driving scene understanding. In: IJCAI. pp. 1743--1749 (2022)

\bibitem{zhou2022learning}
Zhou, K., Yang, J., Loy, C.C., Liu, Z.: Learning to prompt for vision-language models. International Journal of Computer Vision  \textbf{130}(9),  2337--2348 (2022)

\end{thebibliography}

\clearpage
\appendix

{\section*{\Large Supplementary Material}}

The supplementary material is organized as follows: (1) Additional results of Iterations between MIC and MIC with CoDA in Section~\ref{A1}; (2) Ablation studies of severity threshold in~\ref{A2}; (3) Additional ablation studies of CoD in~\ref{A3}; (4) Discussion about Severity-Aware Visual Prompt Tuning in~\ref{B1} including motivation of the Severity metric and visualization of visual prompts; (5) Discussion about the generated dataset in~\ref{B2} containing three rationales and demonstration of images generated with different stages of phrasal prompts.

\section{Additional Results}
\subsection{Additional quantitative and qualitative results.}
\label{A1}
\begin{figure}[t]
    \centering
    \includegraphics[width=\textwidth]{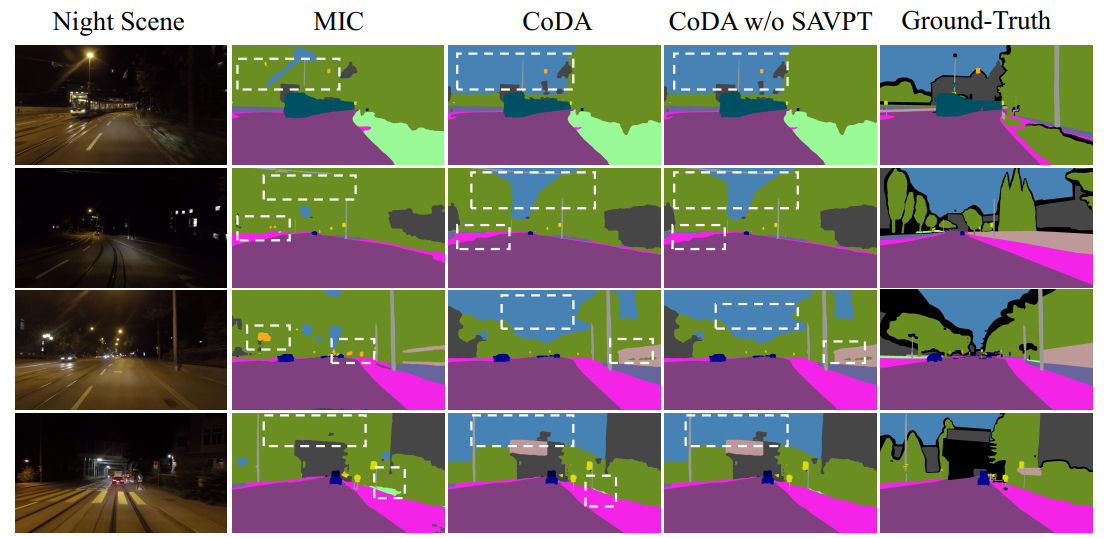}
    \caption{Quantitative experiments between MIC, MIC trained with CoDA, and MIC trained with CoDA but without SAVPT during inference time on ACDC night scene.}
    \label{night}
\end{figure}

\begin{figure}[t]
    \centering
    \includegraphics[width=\textwidth]{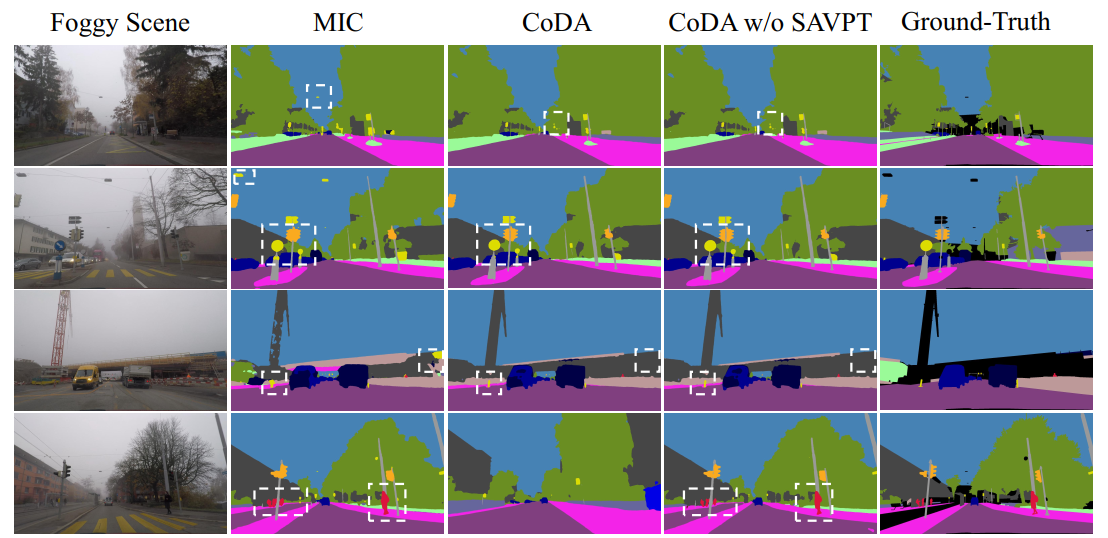}
    \caption{Quantitative experiments between MIC, MIC trained with CoDA, and MIC trained with CoDA but without SAVPT during inference time on ACDC foggy scene.}
    \label{fog}
\end{figure}

\begin{figure}[t]
    \centering
    \includegraphics[width=\textwidth]{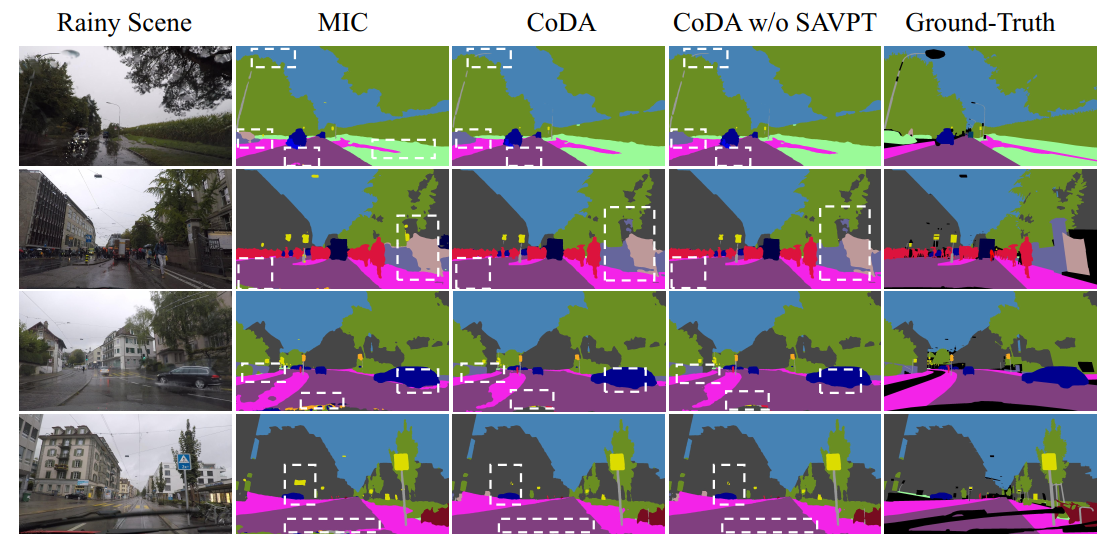}
    \caption{Quantitative experiments between MIC, MIC trained with CoDA, and MIC trained with CoDA but without SAVPT during inference time on ACDC rainy scene.}
    \label{rain}
\end{figure}

\begin{figure}[t]
    \centering
    \includegraphics[width=\textwidth]{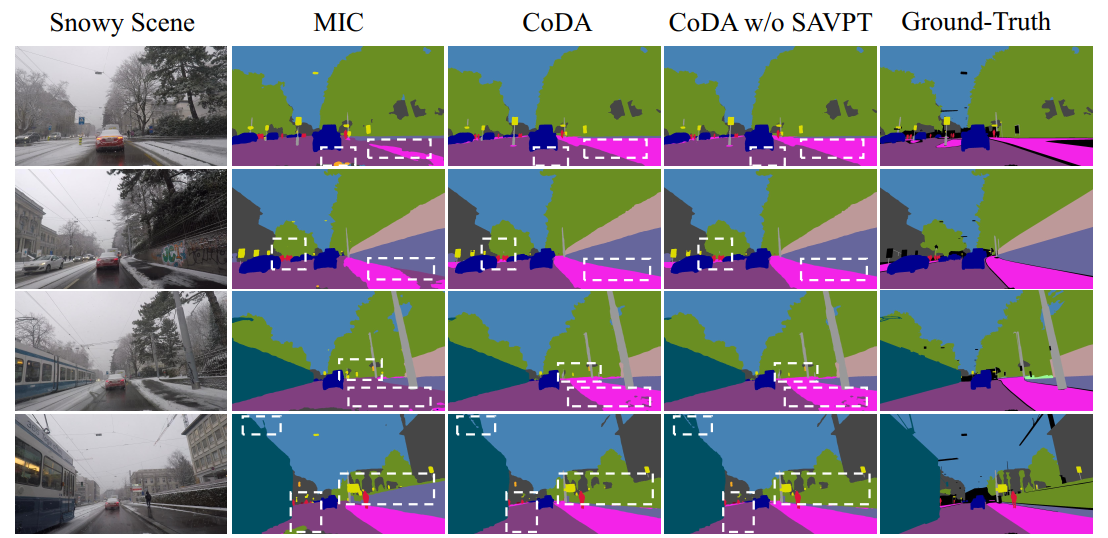}
    \caption{Quantitative experiments between MIC, MIC trained with CoDA, and MIC trained with CoDA but without SAVPT during inference time on ACDC snowy scene.}
    \label{snow}
\end{figure}

We also study the experimental setting of Cityscapes to ACDC, where we train CoDA with 40k and 60k iterations, respectively. The best results of CoDA are trained with 60k Iteration. As shown in Table \ref{iter}, the Origin MIC \cite{hoyer2023mic} underperforms, even with an increased number of iterations, showing a 0.6\% mIoU decline in performance. However, MIC with CoDA outperforms the Origin MIC when trained for both 40k and 60k iterations, thereby demonstrating the impact of MIC with CoDA. Specifically, MIC with CoDA obtains a 0.8\% mIoU improvement when trained for 60k iterations. The results show that CoDA enhances the performance of MIC. We show additional qualitative results under night, fog, rain, and snow adverse scenes in Fig \ref{night}, \ref{fog}, \ref{rain}, and \ref{snow}, respectively.
\begin{table}[htbp]
  \centering
  \caption{Comparison of MIC and MIC trained with CoDA on CS to ACDC val with 40k and 60 iterations.}
    \begin{tabular}{ccc}
    \toprule
    
    \multirow{2}{*}{models} & \multicolumn{2}{c}{Iteration} \\
    \cline{2-3}
                            &\multicolumn{1}{c}{40k} & \multicolumn{1}{c}{60k}\\
    \cmidrule(lr){1-1}\cmidrule{2-3}
    Origin MIC & 68.3 & 67.7\\
    MIC with CoDA & 71.3& 72.1\\
    \bottomrule
    \end{tabular}
  \label{iter}%
\end{table}%
\begin{table}[t]
\centering
\scriptsize
\caption{Ablation studies of CoDA components on CS to ACDC val. VPT means SAVPT without SPT. All mIoU scores are the highest scores during the training.}
\resizebox{0.7\linewidth}{!}{
\begin{tabular}{cccccccc}
\toprule
 Num $\quad$ & MIC $\quad$ & CoD+Tra. $\quad$ & $X_G$ $\quad$ & VPT $\quad$ & SAVPT$\quad$ & mIoU$\quad$ & Gain  \\ 
\midrule
1 $\thinspace$ & \checkmark $\quad$ & -  $\quad$ & -  $\quad$  & -    $\quad$ & - $\quad$ & 68.3  $\quad$ & -     \\
2 $\thinspace$ & \checkmark $\quad$ & -  $\quad$ & - $\quad$   & - $\quad$ & \checkmark $\quad$ & 64.3 $\quad$ & $-4.0$ \\
3 $\thinspace$ & \checkmark $\quad$ & \checkmark  $\quad$ & -  $\quad$  & -   $\quad$ & - $\quad$ & 70.3 $\quad$ & $+2.0$  \\
4 $\thinspace$ & \checkmark $\quad$ & \checkmark  $\quad$ & - $\quad$   & - $\quad$ & \checkmark $\quad$ & 68.4 $\quad$ & $+0.1$ \\
5 $\thinspace$ & \checkmark $\quad$  & \checkmark  $\quad$ & \checkmark  $\quad$  & -   $\quad$ & - $\quad$ & 70.4 $\quad$ & $+2.1$   \\
6 $\thinspace$ & \checkmark $\quad$  & \checkmark $\quad$  & \checkmark  $\quad$  & \checkmark $\quad$ & - $\quad$ & 67.5 $\quad$ & $-0.8$  \\
7 $\thinspace$ & \checkmark $\quad$ & \checkmark  $\quad$ & \checkmark $\quad$   & - $\quad$ & \checkmark $\quad$ & 72.1 $\quad$ & $+3.8$   \\
\bottomrule
\end{tabular}}
\label{table3}
\end{table}

\textbf{Ablation studies of CoDA components}
We conduct experiments to show the necessity of each component in CoDA in Table \ref{table3}. we combine 1, 2, and 4 rows to reveal that the prerequisite of SAVPT is the good foundation brought by the CoD+traditional strategy. From 4, 5, and 7 rows, we know that $X_G$ and SAVPT need to collaborate to maximize the effects with 0.1\% to 3.8\% gain from 4 to 7 rows and 2.1\% to 3.8\% gain from 5 to 7 rows. Combining 6 and 7 rows, results show that SPT is decisive for CoDA's performance bringing 4.2\% improvements from 67.5 to 72.1 mIoU.

\subsection{Ablation study of different severity threshold } 
\label{A2}
The \textit{Severity} threshold $\tau$ enforces models to learn domain-invariant features by focusing on unified severities so that models can ignore the scene-specific features.
For each scene benchmark, we adopt different $\tau$ settings. We show the ablation study on Nighttime Driving validation set in Table \ref{spt}. The results demonstrate that $\tau$=0.05 achieves the best score and outperforms the SOTA method by 0.6\% mIoU.

\renewcommand{\arraystretch}{1.2} 
 \begin{table}[htbp]
\centering
\scriptsize
\caption{Ablation study of CoDA components on the CS to ACDC validation set. Each of the reported mIoU scores (bottom row) is the highest score during the training. The best result is in bold, and the second best is underlined.}
\begin{tabular}{ccccccccccc}
\toprule
\multicolumn{11}{c}{Nighttime Driving Val} \\
\cmidrule(lr){1-1}\cmidrule{2-11}
     $\tau$ & 0.03 & 0.05 &  0.06 &  0.07 & 0.08 & 0.1 & 0.15 & 0.2 & 0.3 & 0.38  \\ 
     CoDA & 56 & \textbf{59.2} & 55 & 55.2 & 53.9 & 54.2 & 56.5 & 56.4 & \underline{57.1} & 55.9\\
\bottomrule
\end{tabular}
\label{spt}
\end{table}

\subsection{Ablation study of Creation of Chain-of-Domain Strategy}
\label{A3}
The training iteration of CoD is critical and we show these experiments in Table \ref{table3}. Although there are 2 intermediate domains in CoD designs, we ablate four intermediate domain settings with different iterations. In Table \ref{table3}, we first try three experiments: 2, 3, and 4 with four intermediate domains (rain, fog, snow, and night) comprising 300, 400, and 1000 iterations, respectively. We observe that the setting with 400 iterations is better than both the 300 and 1000 ones. Since we argue that the most challenging night scene needs to be treated more carefully, we begin by increasing the training iterations of the night scene to 4k. We choose the rain, fog, and snow as $\mathcal{M}_\textit{1}$, $\mathcal{M}_\textit{2}$, and $\mathcal{M}_\textit{3}$, respectively after observing that 
the performances of the previous methods \cite{gong2023train, ma2022both,hoyer2022daformer,hoyer2023mic,hoyer2022hrda} can be ranked from high to low with rain $>$ fog $>$ snow $>$ night. Also, rain, fog, and snow benchmarks report almost equal performances. We, therefore, argue that the rain, fog, and snow should be composited to a unified intermediate domain distinguished from the one comprising night scenes. Combining experiments 3 and 5, we thereby conduct experiments 6, 7, and 8; and conclude the strategy used in the paper. Notably, since we maintain that CoD should be a simple strategy, we do not conduct too many attempts at the iterations and domain compositions. 
Exploring the impact of a diverse number of iterations and variations in domain compositions within CoD will be a focus for future research endeavors.
\begin{table}[]
\centering
\caption{Ablation study of the creation of Chain-of-Domain Strategy on the Cityscapes to ACDC Validation set. The iteration of 10.4k, 15.6k, and 12k in experiments 6, 7, and 8, respectively, signify the total iterations of the intermediate domain adaptation, namely, $S\rightarrow\mathcal{M}_\textit{1}\rightarrow\mathcal{M}_\textit{2}$. Rain, fog, snow, and night indicate the four adverse scenes. Note that r$+$f$+$s refers to the rain$+$fog$+$snow compositional scene, and `all' means r$+$f$+$s$+$n. The best results are in bold and the second best is underlined.}
\label{table3}
\scriptsize
\begin{tabular}{ccccccccc}
\toprule
\multirow{2}{*}{Experiments}$\enspace$$\enspace$ $\enspace$& \multirow{2}{*}{Domains} & \multicolumn{5}{c}{Iteration of Intermediate domains}        & \multirow{2}{*}{mIoU} &$\enspace$$\enspace$\multirow{2}{*}{Gain} \\
                                  &                         & 300   &$\enspace$ 400   &$\enspace$$\thinspace$ 1000  &$\enspace$$\thinspace$ 1200  &$\thinspace$$\thinspace$ 4000  &                       \\ \cmidrule(lr){1-2}\cmidrule(lr){3-7}\cmidrule(lr){8-9}
1 (MIC)$\enspace$$\enspace$ $\enspace$                           & $T$ (all)                       & -     & -     & -     & -     & -     & 68.3 & $\enspace$ -                 \\ \cmidrule(lr){1-2}\cmidrule(lr){3-7}\cmidrule(lr){8-9}
\multirow{4}{*}{2}$\enspace$$\enspace$ $\enspace$                & $\mathcal{M}_\textit{1}$                      & rain   & -     & -     & -     & -     & \multirow{4}{*}{65.4} &$\enspace$ \multirow{4}{*}{$-$2.9}  \\
                                  & $\mathcal{M}_\textit{2}$                         & fog   & -     & -     & -     & -     &                       \\
                                  & $\mathcal{M}_\textit{3}$                          & snow  & -     & -     & -     & -     &                       \\
                                  & $\mathcal{M}_\textit{4}$                          & night & -     & -     & -     & -     &                       \\ \cmidrule(lr){1-2}\cmidrule(lr){3-7}\cmidrule(lr){8-9}
\multirow{4}{*}{3}$\enspace$$\enspace$ $\enspace$                & $\mathcal{M}_\textit{1}$                          & -     & rain  & -     & -     & -     & \multirow{4}{*}{67} &$\enspace$ \multirow{4}{*}{$-$1.3}    \\
                                  & $\mathcal{M}_\textit{2}$                          & -     & fog   & -     & -     & -     &                       \\
                                  & $\mathcal{M}_\textit{3}$                          & -     & snow  & -     & -     & -     &                       \\
                                  & $\mathcal{M}_\textit{4}$                          & -     & night & -     & -     & -     &                       \\ \cmidrule(lr){1-2}\cmidrule(lr){3-7}\cmidrule(lr){8-9}
\multirow{4}{*}{4}$\enspace$$\enspace$ $\enspace$                & $\mathcal{M}_\textit{1}$                          & -     & -     & rain  & -     & -     & \multirow{4}{*}{65.9} &$\enspace$ \multirow{4}{*}{$-$2.4} \\
                                  & $\mathcal{M}_\textit{2}$                          & -     & -     & fog   & -     & -     &                       \\
                                  & $\mathcal{M}_\textit{3}$                          & -     & -     & snow  & -     & -     &                       \\
                                  & $\mathcal{M}_\textit{4}$                         & -     & -     & night & -     & -     &                       \\ \cmidrule(lr){1-2}\cmidrule(lr){3-7}\cmidrule(lr){8-9}
\multirow{4}{*}{5}$\enspace$$\enspace$ $\enspace$                & $\mathcal{M}_\textit{1}$                         & -     & rain  & -     & -     & -     & \multirow{4}{*}{67.1} &$\enspace$ \multirow{4}{*}{$-$1.2} \\
                                  & $\mathcal{M}_\textit{2}$                          & -     & fog   & -     & -     & -     &                       \\
                                  & $\mathcal{M}_\textit{3}$                          & -     & snow  & -     & -     & -     &                       \\
                                  & $\mathcal{M}_\textit{4}$                          & -     & -     & -     & -     & night &                       \\ \cmidrule(lr){1-2}\cmidrule(lr){3-7}\cmidrule(lr){8-9}
\multirow{3}{*}{6 (10.4k)}$\enspace$$\enspace$ $\enspace$        & $\mathcal{M}_\textit{1}$                          & -     & -     & -     & r+f+s & -     & \multirow{3}{*}{69.2} &$\enspace$ \multirow{3}{*}{$+$0.9} \\
                                  & $\mathcal{M}_\textit{2}$                          & -     & -     & -     & -     & night &                       \\
                                  & $T$ (all)                 & -     & -     & -     & -     & -     &                       \\ \cmidrule(lr){1-2}\cmidrule(lr){3-7}\cmidrule(lr){8-9}
\multirow{3}{*}{7 (15.6k)}$\enspace$$\enspace$ $\enspace$        & $\mathcal{M}_\textit{1}$                         & -     & -     & -     & r+f+s & -     & \multirow{3}{*}{\underline{70.1}} &$\enspace$ \multirow{3}{*}{\underline{$+$1.8}} \\
                                  & $\mathcal{M}_\textit{2}$                         & -     & -     & -     & -     & night &                       \\
                                  & $T$ (all)                 & -     & -     & -     & -     & -     &                       \\ \cmidrule(lr){1-2}\cmidrule(lr){3-7}\cmidrule(lr){8-9}
\multirow{3}{*}{8 (12k)}$\enspace$$\enspace$ $\enspace$        & $\mathcal{M}_\textit{1}$                          & -     & -     & -     & r+f+s & -     & \multirow{3}{*}{\textbf{70.3}} &$\enspace$ \multirow{3}{*}{\textbf{$+$2.0}}  \\
                                  & $\mathcal{M}_\textit{2}$                          & -     & -     & -     & -     & night &                       \\
                                  & $T$ (all)                 & -     & -     & -     & -     & -     &                       \\ \bottomrule
\end{tabular}
\end{table}

\textbf{Performance on clean scene benchmark} Additionally, we test CoDA on clean scene benchmark images from GTA\large\romannumeral5\small~\cite{richter2016playing} to Cityscapes \cite{cordts2016cityscapes}. Since there are almost no discrepancies between images with clean weather conditions, we do not use the CoD strategy and Severity Perception Trigger in this experiment. We only activate single-branch Meta-Visual Prompts with Meta-adapters for simply testing the generalizability of these two components. The results are shown in Table \ref{gta} and CoDA* achieves 1.1\% improvement under clean weather conditions.

\begin{table}[htbp]
  \centering
  \caption{Comparison on clean scene images from Cityscapes benchmark. CoDA* represents the results with CoDA's single-branch Meta-Visual Prompts and Meta-Adapters. This experiment is only used for testing the generalizability of the Meta-Visual Prompts and Meta-Adapters components of SAVPT. The best results are in bold and the second best is underlined.}
  \resizebox{\textwidth}{!}{
    \begin{tabular}{ccccccccccccccccccccc}
    \toprule
    Method & Road & S.walk\newline{} & Build & Wall & Fence & Pole & T.Light & Sign & Veget & Terrain\newline{} & Sky & Person & Rider & Car & Truck & Bus & Train & M.bike & Bike & mIoU \\
    \cmidrule(lr){1-1}\cmidrule(lr){2-20}\cmidrule(lr){21-21}
    \multicolumn{21}{c}{GTA\large\romannumeral5\small $\enspace$to Cityscapes} \\
    \cmidrule(lr){1-1}\cmidrule(lr){2-20}\cmidrule(lr){21-21}
    DAFormer&95.7&70.2&89.4&53.5&48.1&49.6&55.8&59.4&89.9&47.9&92.5&72.2&44.7&92.3&74.5&78.2&65.1&55.9&61.8&68.3\\
    HRDA&96.4&74.4&91.0&\underline{80.1}&51.5&57.1&63.9&69.3&91.3&48.4&94.2&79&52.9&93.9&84.1&85.7&75.9&63.9&67.5&73.8\\
    MIC &\underline{97.4}&\underline{80.1}&\underline{91.7}&61.2&\underline{56.9}&\underline{59.7}&\underline{66.0}&\underline{71.3}&\underline{91.7}&\underline{51.4}&\underline{94.3}&\underline{79.8}&\underline{56.1}&\underline{94.6}&\underline{85.4}&\underline{90.3}&\underline{80.4}&\textbf{64.5}&\underline{68.5}&\underline{75.9} \\
    CoDA*& \textbf{97.7}&\textbf{81.6}&\textbf{91.9}&\textbf{63.6}&\textbf{60.6}&\textbf{61.0}&\textbf{67.5}&\textbf{74.0}&\textbf{91.7}&\textbf{52.0}&\textbf{94.7}&\textbf{80.9}&\textbf{58}&\textbf{95}&\textbf{85.9}&\textbf{91}&\textbf{82.2}&\underline{63.7}&\textbf{69.1}&\textbf{77.0}\\

    \cmidrule(lr){1-1}\cmidrule(lr){2-20}\cmidrule(lr){21-21}
    \end{tabular}}
  \label{gta}
\end{table}

\section{Discussion}
\subsection{About Severity-Aware Visual Prompt Tuning (SAVPT)}
\label{B1}
\begin{figure}[t]
    \centering
    \includegraphics[width=\textwidth]{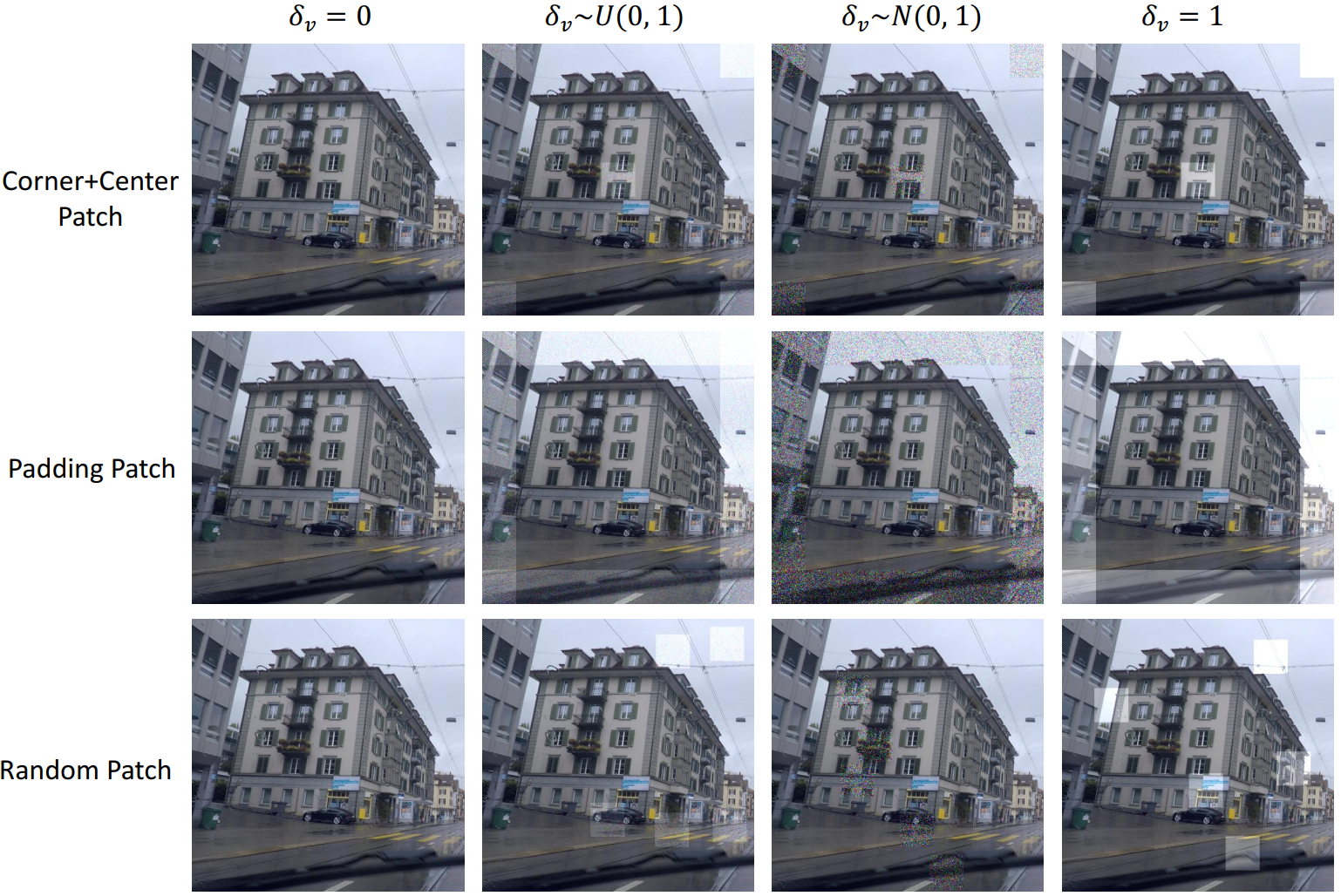}
    \caption{Visualizations of three placements with four initial parameters of Meta-Visual Prompts. Corner+ Center Patch shows Meta-visual prompts are placed on the corner and center positions of images. Padding Patch means Meta-Visual Prompts pad around the images and Random Path demonstrates that five visual prompt blocks are placed on the random positions of images. Best seen on screen and when zoomed-in.}
    \label{vp}
\end{figure}

In the main paper, we discussed the ablation studies of the placement and initial parameters of Meta-Visual Prompts. In Fig \ref{vp}, we show the illustrations of those experiments to validate our results.

\textbf{Motivation Behind the Severity Metric}
Our formulation of the severity metric, denoted by the threshold $\tau$, is intrinsically linked to the concept of image darkness, which also serves as the foundational motivation for our approach. As discussed in the main text, conventional methods~\cite{hoyer2022daformer,gong2023train,hoyer2022hrda,hoyer2023mic} encounter significant challenges with night scene benchmarks, primarily due to the reduced illumination characteristic of night scenes. Darkness, a prevalent feature across various adverse conditions, is particularly pronounced in the most demanding scenarios. This observation leads us to posit that darkness can be seen as a representation of the complexity of recognition tasks, a notion we term \textit{Severity}. Consequently, we propose the adoption of this darkness-centric \textit{Severity} metric as a unified measure for all adverse conditions. In this way, models can be trained by SAVPT to focus on common features to extract domain-invariant features rather than scene-specific features.


\subsection{The Generated Dataset}
\label{B2}
\textbf{Rationale for Incorporating an Additional Dataset:} The ACDC dataset \cite{sakaridis2021acdc} serves as a primary reference in our experiments. Despite categorizing all daytime images as \textit{easy} within our CoD framework, we observe that a significant portion of these images presents considerable challenges for model learning during the initial training phases. To address this, we have curated our own dataset comprising images under slightly altered weather conditions with mild adverse weather conditions, aimed at providing simpler training instances. Moreover, the variance between the daytime images from ACDC and our synthesized images is minimal, preventing the necessity of classifying them into distinct domains. Consequently, we integrate the daytime images from ACDC with our generated images to construct $X_{\mathcal{M}_1}$. While the ACDC dataset includes additional clean weather images that are captured from the same positions as the adverse scene images, our initial attempts to utilize these as intermediary domain images yielded suboptimal results. We argue that the transition from clear to adverse conditions introduces excessive disparity. Models transitioning from a clean source domain through a clean intermediary to an adverse target domain experience increased discrepancies. This led us to the decision of generating images that exhibit mildly adverse weather conditions, to mitigate such disparities.

\textbf{Rationale behind generating phrasal prompts before creating our methodology:} Prior to dataset creation, we experimented with employing both phrasal and sentential prompts to guide the image generation process in Stable Diffusion 2 (SD2) \cite{rombach2022high}. Despite these prompts conveying analogous semantic content, our observations led to the conclusion that SD2 exhibits a better comprehension of phrasal prompts, whereas its capability to parse and interpret long sentences is comparatively limited. Inspired by the methodologies delineated in Chain-of-Thought (CoT) related research \cite{kojima2022large}, we posit that a stepwise instruction generation approach, as facilitated by GPT-4V, yields a more focused approach. Consequently, we have opted to integrate a series of progressively optimized phrasal prompts within our pipeline. This strategy is designed to enhance SD2's interpretative accuracy, thereby facilitating the production of images of superior quality.


\textbf{Rationale behind choosing several images for prompt generation in stage 1:} 
The primary objective of our dataset is to serve as a bridge between the source and target domains. We, therefore, need to obtain prompts that can represent these two domains. We argue that these common prompts should demonstrate the basic features of these domains, thereby enhancing the image generation process's robustness. Consequently, the prompts generated during the initial phase are designed to reflect basic urban features, such as \textit{hazy street views} and \textit{pedestrian crossings}, as opposed to more detailed and precise features, like \textit{a human standing next to a Benz car}. 
Meanwhile, each image generates similar prompts with redundant information, so we randomly choose two images from each scene. The results in the main paper also show that our generated images indeed make differences in the adaptation process. 
\begin{figure}[t]
    \centering
    \includegraphics[width=\textwidth]{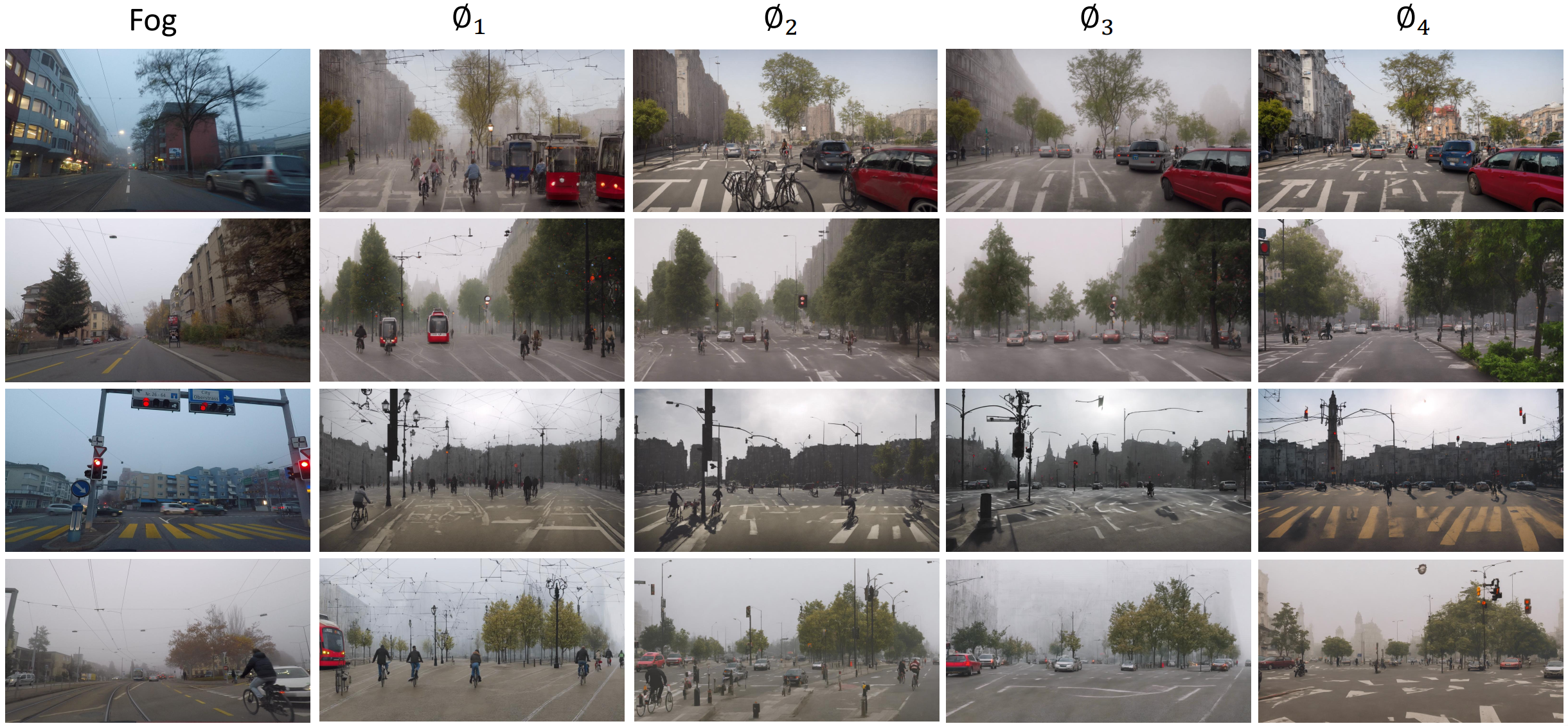}
    \caption{Quantitative experiments of foggy images' qualities generated by different stage prompts ($\Phi_1$, $\Phi_2$, $\Phi_3$, and $\Phi_4$).}
    \label{fog_generate}
\end{figure}

\textbf{Generated images with different stages of phrasal prompts} 
We systematically iterate through a four-stage prompt generation process: $\Phi_1$ encompasses the initial phrasal prompts that delineate the features of the target domains; $\Phi_2$ consists of carefully curated prompts that encapsulate the shared and basic features of both the source and target domains; $\Phi_3$ comprises prompts refined by GPT-4V, building upon the foundation laid by $\Phi_2$; and $\Phi_4$ represents the final prompts, crafted for the final image-to-image translation tasks.

\begin{figure}[t]
    \centering
    \includegraphics[width=\textwidth]{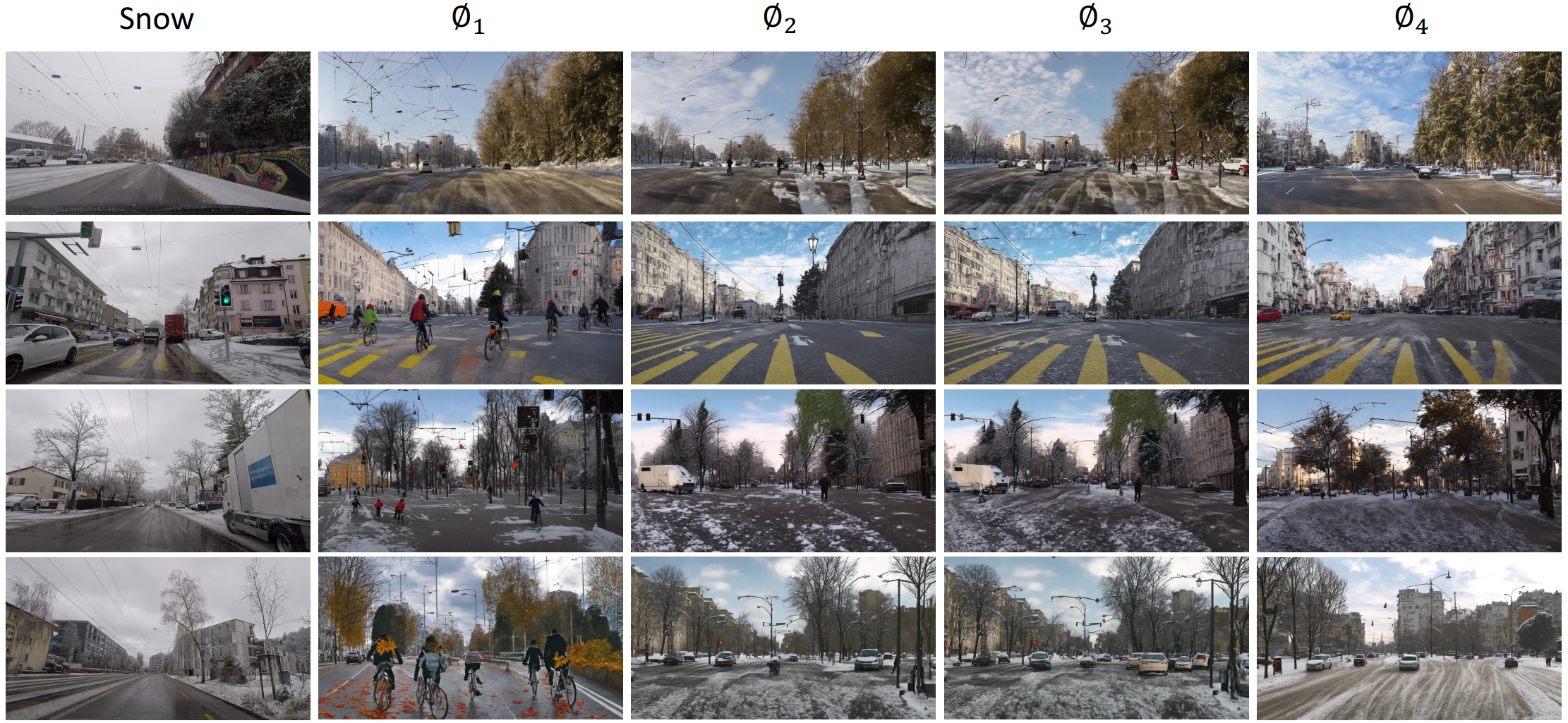}
    \caption{Quantitative experiments of snowy images' qualities generated by different stage prompts ($\Phi_1$, $\Phi_2$, $\Phi_3$, and $\Phi_4$).}
    \label{snow_generate}
\end{figure}

\begin{figure}
    \centering
    \includegraphics[width=\textwidth]{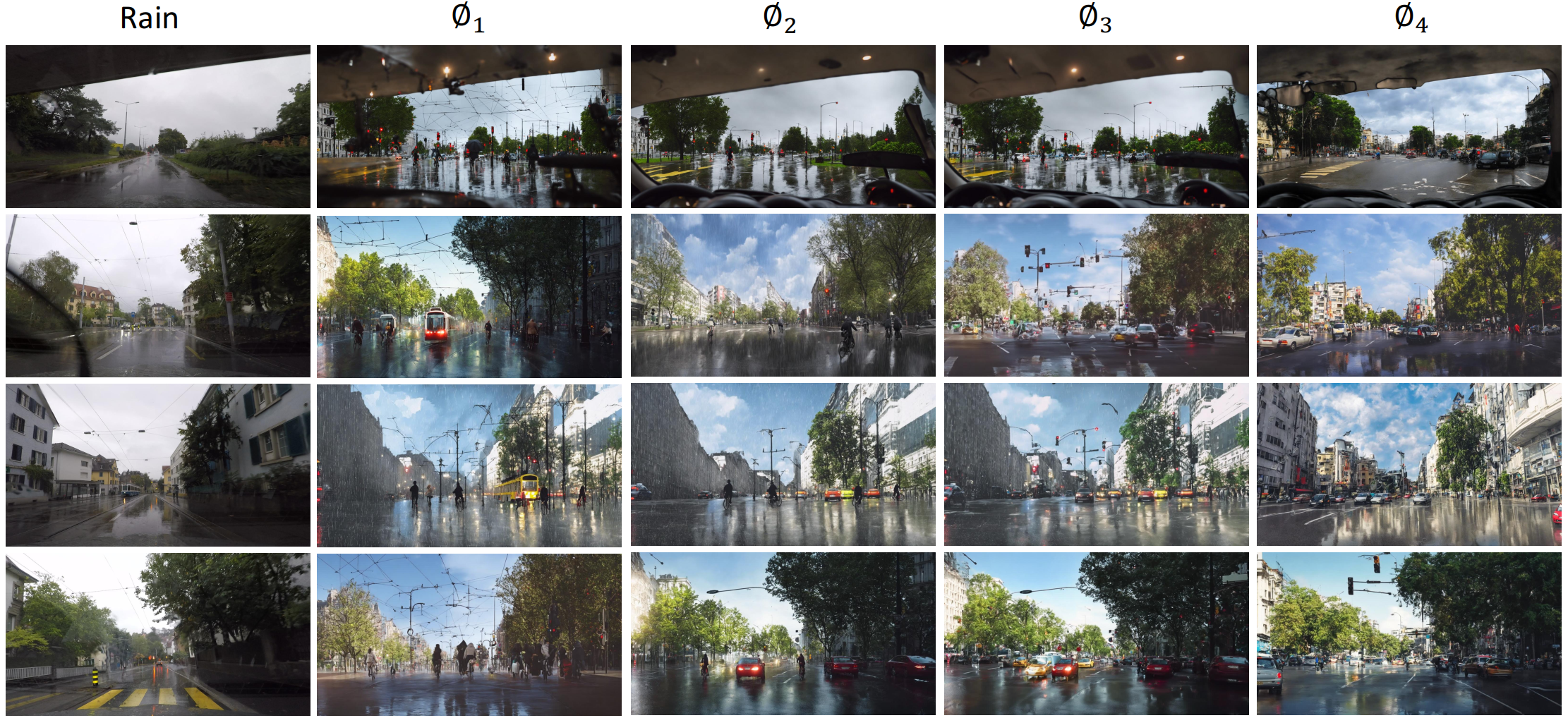}
    \caption{Quantitative experiments of rainy images' qualities generated by phrasal prompts in different stages ($\Phi_1$, $\Phi_2$, $\Phi_3$, and $\Phi_4$).}
    \label{rain_generate}
\end{figure}

Visual representations of images depicting fog, snow, and rain, accompanied by the corresponding four stages of phrasal prompts $\Phi$, are depicted in Figures \ref{fog_generate}, \ref{snow_generate}, and \ref{rain_generate}, respectively. In the initial stage of our experiments, we generate $\Phi_1$ randomly for each scenario by employing phrasal weather prompts, such as "<rainy/snowy/foggy> street view," alongside four randomly chosen prompts from $\Phi_1$ across all scenes. For subsequent stages of generation, we adhere to the progression from $\Phi_2$ through $\Phi_4$, with edits limited only to the weather conditions depicted in the prompts. 
Upon integrating $\Phi_1$ and $\Phi_2$, it becomes evident that images produced using $\Phi_2$ generally demonstrate cleaner and better texture details compared to those generated from $\Phi_1$. Specifically, the sky generated in images from $\Phi_2$ surpasses the quality of those from random $\Phi_1$, notably avoiding the erroneous inclusion of traffic lights in the sky. Furthermore, the $\Phi_2$ generated images are characterized by a reduction in anomalous objects, such as low-quality depictions of cyclists and trains misplaced on roads, when compared to $\Phi_1$. These findings suggest that $\Phi_2$ prompts, which encapsulate the shared and basic features of both source and target domains, are more beneficial to the performance of Stable Diffusion 2 (SD2) than the random $\Phi_1$ prompts that focus solely on target domain features. This evidence validates the effectiveness of the second stage in our methodology and aligns well with our above motivation that the common and basic features indeed can strengthen the robustness of SD2. 

In the experiments involving $\Phi_3$ and $\Phi_4$, we assess the images against those generated in $\Phi_2$. For foggy and snowy conditions, the images from $\Phi_3$ and $\Phi_4$ not only exhibit semantic enrichment but also improve the clarity of weather conditions. Conversely, in the rainy scenario, images produced with $\Phi_3$ prompts introduce a greater number of objects due to the semantic enhancements implemented in stage 3, albeit resulting in a somewhat cluttered appearance compared to $\Phi_2$. Given the overall enhancement in image quality observed in stage 3, no specific adjustments were made for particular images, and this dataset has proven effective in our ACDC experiments.



\end{document}